\documentclass[conference]{IEEEtran}
\usepackage{ulem}
\usepackage{threeparttable}
\usepackage{multirow}
\let\labelindent\relax
\usepackage{enumitem}
\usepackage{caption}
\usepackage{color}
\usepackage[linesnumbered,ruled]{algorithm2e}
\makeatletter
\renewcommand{\@algocf@capt@plain}{above}
\makeatother 
\usepackage{algorithmicx}  
\usepackage{algpseudocode}

\usepackage{amsmath,amsthm} 
\algnewcommand{\LineComment}[1]{\State \(\triangleright\) #1}
\usepackage{soul}
\usepackage{makecell}
\usepackage{subfigure}
\usepackage{breqn}
\usepackage{mathtools}
\usepackage[backref]{hyperref} 
\hypersetup{
hidelinks}

\ifCLASSINFOpdf
\else
\fi
\begin{document}
%
\title{PPA: Preference Profiling Attack \\ Against Federated Learning}

\author{
   \IEEEauthorblockN{Chunyi Zhou\IEEEauthorrefmark{2}\IEEEauthorrefmark{4}\IEEEauthorrefmark{1}, Yansong Gao\IEEEauthorrefmark{2}\IEEEauthorrefmark{1}, \\ Anmin Fu\IEEEauthorrefmark{2}\IEEEauthorrefmark{4}\IEEEauthorrefmark{3}, Kai Chen\IEEEauthorrefmark{4}, Zhiyang Dai\IEEEauthorrefmark{2}, Zhi Zhang\IEEEauthorrefmark{5}, Minhui Xue\IEEEauthorrefmark{5}, and Yuqing Zhang\IEEEauthorrefmark{6}}

  \IEEEauthorblockA{\textit{\IEEEauthorrefmark{2}School of Computer Science and Engineering, Nanjing University of Science and Technology}, China 
  }
  \IEEEauthorblockA{\textit{\IEEEauthorrefmark{4}State Key Laboratory of Information Security, Institute of Information Engineering, Chinese Academy of Science}, China}
  \IEEEauthorblockA{\textit{\IEEEauthorrefmark{5}Data61, CSIRO}, Syndey, Australia}
  \IEEEauthorblockA{\textit{\IEEEauthorrefmark{6}National Computer Network Intrusion Protection Center, University of Chinese Academy of Science}, China}
  \IEEEauthorblockA{\{zhouchunyi;yansong.gao;fuam;dzy\}@njust.edu.cn; chenkai@iie.ac.cn;\\ \{zhi.zhang; jason.xue\}@data61.csiro.au; zhangyq@ucas.ac.cn}
\IEEEauthorblockA{\textit{\IEEEauthorrefmark{1}Co-first authors}; \textit{\IEEEauthorrefmark{3}Corresponding author}}
}



\IEEEoverridecommandlockouts
\makeatletter\def\@IEEEpubidpullup{6.5\baselineskip}\makeatother
\IEEEpubid{\parbox{\columnwidth}{
    Network and Distributed System Security (NDSS) Symposium 2023\\
    28 February - 4 March 2023, San Diego, CA, USA\\
    ISBN 1-891562-83-5\\
    https://dx.doi.org/10.14722/ndss.2023.23171\\
    www.ndss-symposium.org
}
\hspace{\columnsep}\makebox[\columnwidth]{}}

\maketitle

\begin{abstract}
Federated learning (FL) trains a global model across a number of decentralized users, each with a local dataset. Compared to traditional centralized learning, FL does not require direct access to local datasets and thus aims to mitigate data privacy concerns. However, data privacy leakage in FL still exists due to inference attacks, including membership inference, property inference, and data inversion.   

In this work, we propose a new type of privacy inference attack, coined Preference Profiling Attack (PPA), that accurately profiles the private preferences of a local user, e.g., most liked (disliked) items from the client's online shopping and most common expressions from the user's selfies. 
In general, PPA can profile top-$k$ (i.e., $k$ = $1, 2, 3$ and $k = 1$ in particular) preferences contingent on the local client (user)'s characteristics. 
Our key insight is that the gradient variation of a local user's model has a distinguishable sensitivity to the sample proportion of a given class, especially the majority (minority) class. By observing a user model's gradient sensitivity to a class, PPA can profile the sample proportion of the class in the user's local dataset, and thus \textit{the user's preference of the class} is exposed. The inherent statistical heterogeneity of FL further facilitates PPA.
We have extensively evaluated the PPA's effectiveness using four datasets (MNIST, CIFAR10, RAF-DB and Products-10K). Our results show that PPA achieves 90\% and 98\% top-$1$ attack accuracy to the MNIST and CIFAR10, respectively. More importantly, in real-world commercial scenarios of shopping (i.e., Products-10K) and social network (i.e., RAF-DB), PPA gains a top-$1$ attack accuracy of 78\% in the former case to infer the most ordered items (i.e., as a commercial competitor), and 88\% in the latter case to infer a victim user's most often facial expressions, e.g., disgusted.
The top-$3$ attack accuracy of RAF-DB and top-$2$ accuracy is up to 88\% and 100\% for the Products-10K and RAF-DB, respectively. We also show that PPA is insensitive to the number of FL's local users (up to 100 we tested) and local training epochs (up to 20 we tested) used by a user. Although existing countermeasures such as dropout and differential privacy protection can lower the PPA's accuracy to some extent, they unavoidably incur notable deterioration to the global model. \textcolor{blue}{The source code is available at \href{https://github.com/PPAattack}{https://github.com/PPAattack}}.
\end{abstract}


%

\section{Introduction}
As traditional centralized deep learning requires aggregating user data into one place, it can be abused to leak user data privacy and breaches the national regulations such as General Data Protection Regulation (GDPR)~\cite{gdpr}, California Privacy Rights Act (CPRA)~\cite{cpra}, and China Data Security Law (CDSL)~\cite{cdsl}. In contrast to centralized learning, Federated learning (FL)~\cite{McMahan2017Communication} voids the requirement by using local user model updates rather than raw user data, significantly mitigating data privacy leaks~\cite{Xu2019Data,yang2019federated,ZuoLZ19,BeguelinWTRPOKB20}. 
As such, FL has become the most popular distributed machine learning technique and empowered a wide range of privacy-sensitive applications, such as smart healthcare, social network and wireless communication~\cite{Zhang2019Survey,Nguyen2019Infocom1,Chunyi,PapernotAEGT17,SinghS0H21,EmotionAnalysis}.

\begin{figure}[h]
	\centering  
	\includegraphics [width = \columnwidth]{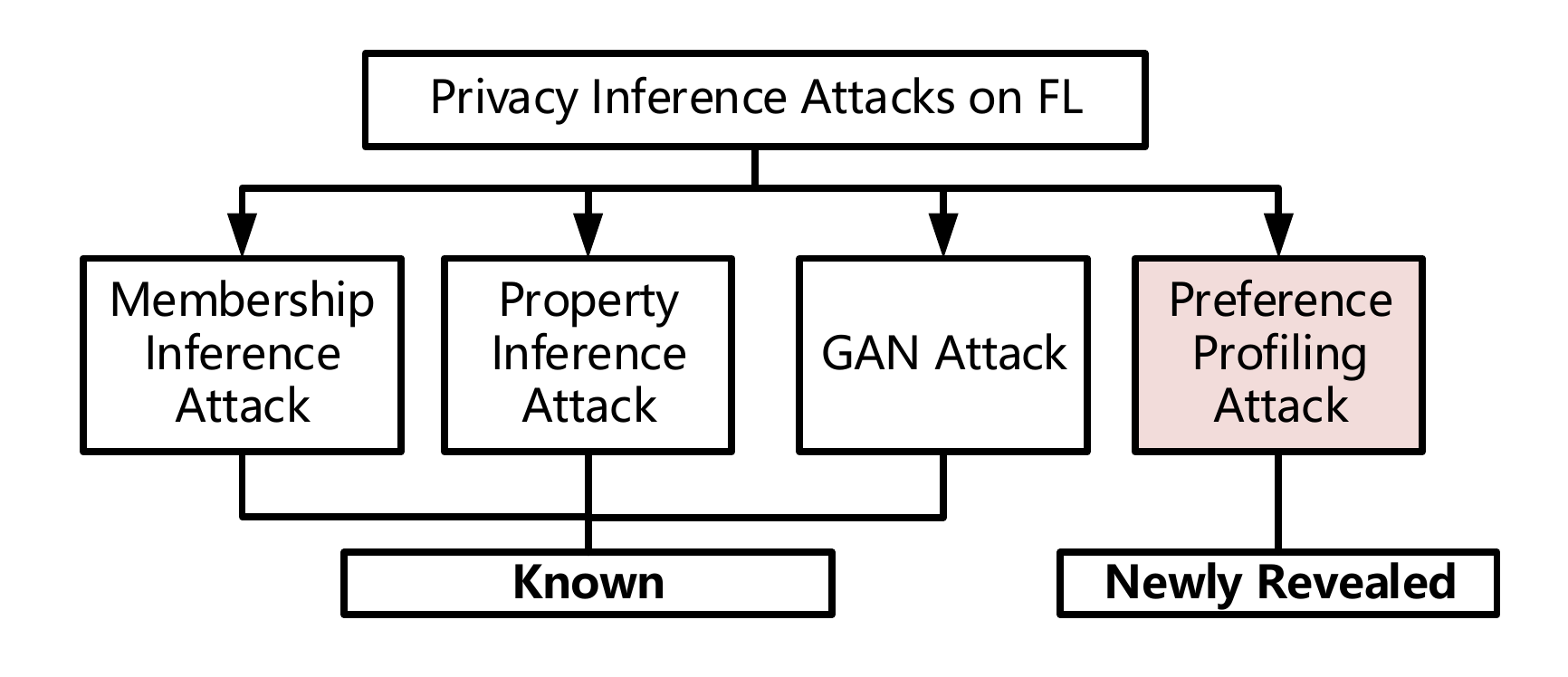}  
	\caption{Privacy inference attacks on FL. 
	}  
	\label{inferenceattack}  
\end{figure} 

\noindent{\bf State-of-the-art Attacks:} Nonetheless, private user information in FL can still be leaked by inference attacks~\cite{Mothukuri2021survey}, as summarized in Figure~\ref{inferenceattack}. 
To date, there are three known privacy inference attacks. Specifically, a membership inference attack~\cite{Shokri2017member,Nasr2019Comprehensive} determines whether a particular data record is used for training a victim user's uploaded model. A property inference attack~\cite{Ganju2018Property,Melis2019Exploiting} infers whether a target attribute exists in a victim user's local dataset. A Generative Adversarial Networks (GAN) attack~\cite{Hitaj2017GAN,Wang2019Beyond} reverse engineers images of a target label based on the model uploaded by a victim user. 
In privacy-sensitive applications, 
these inference attacks can leak private information of a target user, posing severe threats to user privacy in FL~\cite{MayerZSA21}. 
Taking the DNA determination as an example, the membership inference attack can utilize a statistical distance measurement to determine if a known individual is in a mixture of DNA~\cite{HagestedtHBLE0020}.

\textit{None} of the aforementioned privacy inference attacks can profile local user data preference of FL, one of the most sensitive types of personal information an attacker is interested in, akin to profiling users in social networks~\cite{BurayaFFC17,YangC19,AyoraHK21,GuoLCZCZX21}. For instance, in an FL-based recommendation system~\cite{yang2020federated,AnelliDNFN21,QinLQ21}, an attacker would be interested in items liked (disliked) by a user who is a participant in FL, as the best (worst) selling items of a shopping mall is an appealing target for a commercial competitor. 
In this context, we ask the following questions:

\begin{center}
  \textit{Is it feasible to profile user data preference in FL? If so, can we demonstrate an effective preference profiling attack?}
\end{center}

\noindent{\bf Preference Profiling Attack:} We provide affirmative answers and demonstrate that effectively profiling user preference can be achieved by leveraging our presented techniques. A viable PPA attack is built upon our key observation.

\noindent\textit{Key Observation:}
A model memorizes the data-distribution characteristic of a training dataset, and inadvertently reflects it in the form of gradient changes, that is, when the model is being trained upon a dataset, its gradient change is related to the sample size of a class. For example, if data does not exist or has a small amount in the dataset, the model does not have the ability of generalization at the beginning. Thus, the model will exhibit \textit{a greater gradient effect to change the weights of the corresponding neurons to minimize the expected loss of the model}. 
That is, the gradient change or sensitivity in the model training process is inversely proportional to the sample size of a class: a large (small) gradient change or sensitivity will be introduced when the sample size of a class is small (large) (detailed in Section~\ref{sec:rationale}). 
Additionally, FL has two inherent characteristics: system and statistical  heterogeneity~\cite{li2020federated,kairouz2019advances}. The statistical heterogeneity means that FL users have different data distribution in reality, and amplifies the gradient sensitivity discrepancies among diverse classes of the local dataset, which facilitates PPA.

\noindent\textit{Challenges:} To achieve PPA, there are three main challenges:

\begin{itemize}[noitemsep, topsep=2pt, partopsep=0pt,leftmargin=0.4cm]
\item How to extract and quantify the gradient sensitivity of a local model per class (label)?

\item How to improve the precision of the gradient sensitivity in a fine-grained manner?

\item How to profile the sample size proportion of a class label given the quantified sensitivity?
\end{itemize}

\textit{Our Solutions:} For the first challenge, given a user uploaded model, we iteratively retrain it with a few samples per class to extract the gradient sensitivity (variations) per class. A larger sample size proportion of class leads to a lower sensitivity. For the second challenge, we strategically select a subset of users to aggregate a subset of models of interested users for the global model update---this is alike the sampling commonly used in FL. We call it selective aggregation. Because trivially aggregating all models will conceal per user's data characteristics, rendering difficulty of saliently extracting the local model sensitivity in the consecutive rounds. This allows the attacker to gain fine-grained sensitivity information of interested users in the next FL round while not degrading the global model utility, thus still remaining stealthy. As for the last challenge, we leverage an attack model that is a meta-classifier to automatically predict targeted user data preference by feeding the extracted sensitivity information of user uploaded model and aggregated model as inputs. 

\textbf{Contributions:} We have made three main contributions:
\begin{itemize}[noitemsep, topsep=2pt, partopsep=0pt,leftmargin=0.4cm]
\item We reveal a new type of privacy inference attack, Preference Profiling Attack (PPA), on FL. By exploiting the fact that the model memorizes the data preference (i.e., Majority (Minority) class) and leaves traces, we construct a model sensitivity extraction algorithm to determine the gradient change per class. Through the designed meta-classifier, PPA can infer the preference class of the user's local data in FL.
	
\item We design a selective aggregation mechanism to greatly improve the success rate of PPA and alleviate the cancellation effect of intuitive global model aggregation on inference attacks. It aggregates a victim user's model with $x$ models with the highest (lowest) model sensitivity of majority (minority) class, that is, the model sensitivity of $x$ local models to the victim local model's majority (minority) class is opposed to that of the targeted local model. The operation can amplify the \textit{differential model sensitivity}, significantly improving efficacy of the meta-classifier for better attack accuracy.  
	
\item We perform a comprehensive evaluation of PPA on privacy-concerned tasks, in particular, online shopping (i.e., Products-10K) and selfies sharing (i.e., RAF-DB) in addition to the MNIST and CIFAR10 based extensive validations. Experimental results affirm that PPA can accurately infer the top-$k$ (e.g., $k  = 1,  2, 3$) preferences of user datasets in FL
contingent on a local user's characteristics. Our experiments demonstrate that PPA retains high efficiency when we increase the number of users in FL and the number of local training epochs adopted by the user. We evaluate PPA against the non-cryptographic defenses of dropout~\cite{Dropout} and differential privacy~\cite{DifferentialPrivacy}, resulting in the conclusion that PPA is still highly effective if the global model does not allow a notable accuracy drop. 
\end{itemize}

\section{Background and Related Work}
In this section, we first introduce FL as well as its commonly used aggregation mechanism \texttt{FedAvg}, and then briefly describe existing privacy inference attacks.
\subsection{Federated Learning} 
FL obviates the problematic raw data sharing from distributed users.
In FL, each user trains local model and only uploads the model rather than any raw data to the server for computing global model of each new round. \texttt{FedAvg} \cite{McMahan2017Communication} is the most well-known aggregation mechanism adopted by FL. In \texttt{FedAvg}, the global model $\theta _{\rm{agg}}$ in round $t$ is computed by:
\begin{equation}
\theta _{\rm{agg}}^{t} = {\textstyle \sum_{n=1}^{N_{\rm user}}} \frac{D_{n} }{D} \theta _{n}^{t} , 
\end{equation}
where $N_{\rm user}$ users participate in FL, each holding $D_{n}$ data points to train local model $\theta _{n}^{t}$ in round $t$. In addition, \texttt{FedAvg} usually utilizes a parameter $C$: the fraction of users that participate in aggregation at a given round. This is useful considering the `straggler' due to network stability or simply leaving of some users. When the sampling rate $C$ is set to 0.1-0.2, FL reaches a good trade-off between computational efficiency and convergence rate, as demonstrated by~\cite{McMahan2017Communication}. Therefore, for the selective aggregation experiments in our work, we align previous work~\cite{McMahan2017Communication} by normally setting the proportion of participated users at each aggregation (i.e. $x$ = 1--4 if $N_{\rm user}=10$) in order to obtain the best experimental effect.

\subsection{Membership Inference Attack} 

The membership inference attack~\cite{Shokri2017member} proposed by Shokri \textit{et al.} constructs shadow models by imitating the behavior of target model, and then trains the attack model according to their outputs, which can infer the existence of a specific data record in the training set. 
Salem \textit{et al.}~\cite{Salem2019} optimized the attack by decreasing the number of shadow models from \textit{n} to 1. Nasr \textit{et al.} \cite{Nasr2019Comprehensive} designed a white-box membership inference attack against centralized and FL by exploiting the vulnerability of stochastic gradient descent algorithm. 
Zari \textit{et al.} \cite{abs-2111-00430} also demonstrated the passive membership inference attack in 
FL. 
Chen \textit{et al.} \cite{Chen2020GAN} provided a generic membership inference attack to attack the deep generative models and judged whether the image belongs to the victim's training set by devising a calibration technique. 
Leino \textit{et al.} \cite{Leino2020Stolen} utilized the model overfitting impact to design a white-box membership inference attack, and demonstrated that this attack outperforms prior black-box methods. Pyrgelis \textit{et al.} \cite{Pyrgelis2018Knock} focused on the feasibility of membership inference attacks on aggregate location time-series, and used adversarial tasks based on game theory to infer membership information on location information. Some membership inference attacks~\cite{Representation2021,HayesMDC19,HilprechtHB19} attacked generative model under the white-box and black-box settings.

\subsection{Property Inference Attack}

Ganju \textit{et al.} \cite{Ganju2018Property} noted that fully connected neural networks are invariant under permutation of nodes in each layer and developed a property inference attack that can extract property information from the model. Melis \textit{et al.} \cite{Melis2019Exploiting} devised a feature leakage attack in collaborative learning, which can infer properties that are independent of the joint model's target properties through holding a subset of training data. Recently, Mathias \textit{et al.} \cite{Mathias2021Property} studied the impact of model complexity on property inference attacks in convolutional neural networks and the results demonstrated that the privacy leakage risk exists independently of the complexity of the target model. Gopinath \textit{et al.}~\cite{Gopinath2019Property} proposed a property inference attack that automatically infers formal properties of feed-forward neural networks. They used encoding convex predicates on the input space to extract the input properties.

\subsection{GAN Attack}

Hitaj \textit{et al.} \cite{Hitaj2017GAN} first proposed GAN attack against FL. The attacker disguises as a normal user to join the model training and obtains the imitation data of other users based on GAN to reconstruct prototypical images of the targeted training set. In addition, Wang \textit{et al.} \cite{Wang2019Beyond} proposed an inference attack based on GAN and multitask discriminator in FL, which achieves user privacy disclosure and is able to recover the private data of the target user on the server-side without interfering with the training process.

\subsection{Privacy Inference Attacks on FL}

In FL, there have been great efforts to explore its privacy leakage through various inference attacks, including membership inference~\cite{Nasr2019Comprehensive, abs-2111-00430}, property inference~\cite{Melis2019Exploiting} and GAN~\cite{Hitaj2017GAN,Wang2019Beyond}. These inference attacks can obtain a variety of user privacy information from the local model. However, none of these attacks can infer the \textit{preference classes} of user dataset. Therefore, our PPA is positioned as a new type of privacy inference attack orthogonal to known ones, as summarized in Figure~\ref{inferenceattack}.

\section{Preference Profiling Attack}\label{sec:related}

In this section, we first define PPA's threat model and clarify its attack goals, then present an overview of PPA, followed by PPA implementation details of each component. To ease the following description and understanding, we mainly use majority class preference profiling for descriptions. But the attack is equally applicable to the minority class preference and further extendable to top-$k$ classes preference profiling.

\begin{figure}
	\centering  
	\includegraphics [width = 0.45\textwidth]{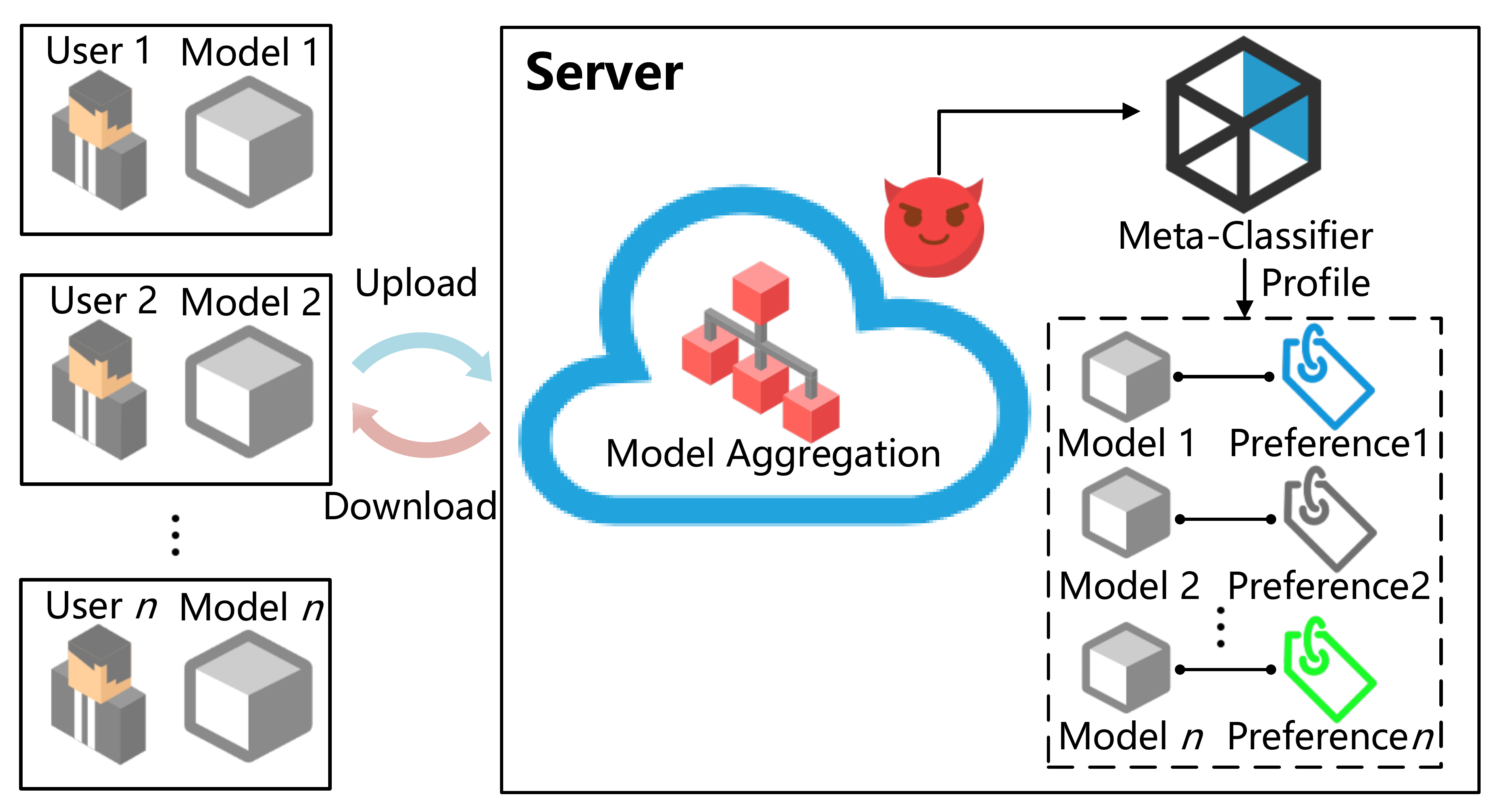}  
	\caption{Threat model of PPA.}  
	\label{Threatmodel}  
\end{figure}

\subsection{Threat Model and Attack Goals}
\subsubsection{\textbf{Threat Model}}
There are two types of entities in FL: local users and the server serving as FL coordinator. Notably, the user here is unnecessarily a human user, it could be a party or an institution e.g., a shopping mall or a hospital. Generally, the server is assumed to be malicious and launches PPA to profile a targeted user's preference through observing the victim's uploaded model, shown in Figure~\ref{Threatmodel}. PPA profiles the user preference class in the plaintext domain---server aggregates the local models in plaintext---rather than in the ciphertext domain, which is aligned with existing privacy inference attack counterparts \cite{Shokri2017member,Nasr2019Comprehensive,Ganju2018Property,Melis2019Exploiting,Hitaj2017GAN} as summarized in Figure~\ref{inferenceattack}.
The capabilities and knowledge of both entities are elaborated as follows:

\begin{itemize}[noitemsep, topsep=2pt, partopsep=0pt,leftmargin=0.4cm]
	\item \textbf{Victims (Users):} Victims are the users whose private local data preferences are interested by the server or attacker.
	Local users participate in FL and collaboratively contribute to learning a global model without directly sharing their locally resided data. Due to statistical heterogeneity, they have different data volumes per class, thus forming diverse data distributions. We assume that users neither trust other users nor share local data and model parameters with each other. Users genuinely follow the FL's procedure and they may employ some common privacy-enhancement techniques, particularly differential privacy and dropout.
	
	\item \textbf{Attacker (Server):} The server intends to profile a victim's preference, especially, majority information from her uploaded model, e.g., shops train the federated model based on users' shopping records, thus using PPA to analyze users' shopping habits. FL enforces privacy protection by local training instead of uploading sensitive data to the server. Therefore, in most inference attacks on FL~\cite{Nasr2019Comprehensive,Melis2019Exploiting,Hitaj2017GAN}, the server usually acts as an attacker trying to infer privacy information from the local model.
	We assume that the server only knows types of users' training datasets and can obtain a small set of benign samples covering all categories, e.g., from public sources, as an auxiliary dataset. The auxiliary dataset does not intersect with any user's training dataset. The assumption of accessible small public dataset in FL is aligned with~\cite{Nasr2019Comprehensive,zhao2018federated,CaoF0G21,DongCLWZ21}.
We assume that the server purposely chooses (samples) a specific subset of local models to update the global model through \textit{model aggregation}, which is a common FL technique to mitigate adverse effects from `straggler' users due to, e.g., unstable network or simply leaving~\cite{li2020federated,li2019convergence}.

\end{itemize}

\subsubsection{\textbf{Attack Goals}}
Overall, PPA is a white-box attack, and the attacker can profile preferences of (all) participated users simultaneously to achieve the following goals:

\begin{itemize}[noitemsep, topsep=2pt, partopsep=0pt,leftmargin=0.4cm]
	\item \textbf{Attack Efficacy.} PPA is to achieve a high attack success rate simultaneously for (all) users in FL, even if the user applies common privacy-preserving mechanisms (i.e., differential privacy and dropout). 

	\item \textbf{Attack Stealthiness.}	This is to ensure that the global model availability and utility are not degraded and its accuracy is on par with that trained via a normal FL without PPA. Since the selective aggregation mechanism we designed does not forge models and disguised under the common sampling strategy, it is difficult for users to be aware that their local models are under attack as the model utility is not affected.
	
	\item \textbf{Attack Generalization.}  This goal is to ensure that our attack is generic to diverse FL settings (e.g., heterogeneous data), and varying preference profiling interests (e.g., most liked, most disliked, top-$k$ with $k$ no less than 1), number of FL users, and number of local training epochs chosen by individual users.
\end{itemize}

\subsection{PPA Overview}
\begin{figure*}
	\centering  
	\includegraphics [width = 0.80\textwidth]{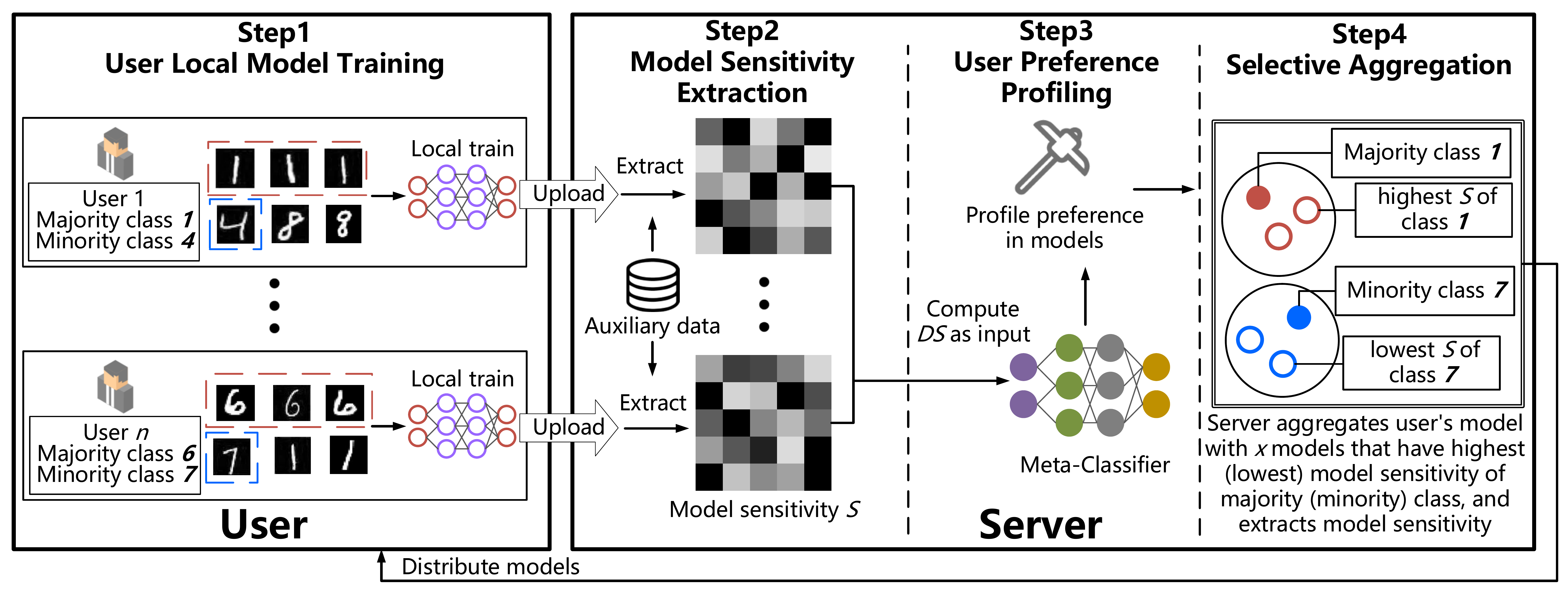}  
	\caption{An overview of PPA.}  
	\label{scheme}  
\end{figure*}

PPA is based on the key observation that sample proportion size of classes in the training dataset can have a direct impact on a model's gradient change sensitivity. The server as an attacker exploits user model sensitivity extraction to profile user (data) preferred class(es), especially majority and minority class that to a large extent is most interested in commercial applications as we later evaluated. 
At the same time, the global model accessed by the user is provided without utility degradation while facilitating PPA by selectively aggregating a subset of user models, namely selective aggregation. Overall, PPA has four steps as illustrated in Figure~\ref{scheme}, each of which is succinctly described below and then elaborated. The notations are summarized in Table \ref{Notation}.

\begin{enumerate}[noitemsep, topsep=2pt, partopsep=0pt,leftmargin=0.4cm]

	\item \textbf{User Local Model Training.}  Users with heterogeneous data train models locally and upload them to the server. 
	
	\item \textbf{Model Sensitivity Extraction.}  After receiving a user local model, the server uses the auxiliary dataset to retrain the model in order to extract the model sensitivity per class.
	
	\item \textbf{User Preference Profiling.}  When getting model sensitivity, the meta-classifier is used to predict a user preferred class(es).  
	
	\item \textbf{Selective Aggregation.} Instead of forming the global model through all user models, for a targeted local model, only $x$ models with the highest (lowest) model sensitivity of its majority (minority) class compared to the targeted model are used in the aggregation. And then this aggregated model is sent to the targeted user---each user may receive a different aggregated model. This is to improve the PPA's attack accuracy. 
\end{enumerate}

Step 4) requires a meta-classifier, which is prepared during the offline phase. In other words, before FL begins among multiple users, the server trains the meta classifier offline based on auxiliary dataset $D _{\rm{aux}}$. 	
The implementation details of each step and the offline meta-classifier training are elaborated in the following. According to the time order, we start by meta-classifier offline training.

\begin{table}
\small
	\caption{Notation summary.}   
	\label{Notation}
	\centering
	\begin{tabular}{p{1.2cm}|p{6.3cm}}   
		\hline  
		\textbf{Notation} & \textbf{Description}  \\   
		\hline  
		$N_{\rm user}$   & Number of users in FL.\\  
		\hline  
		${{\rm{mc}}_n}$  & Majority (minority) class in $n_{\rm th}$ user's local data.\\  
		\hline  
		\textit{T} & Training rounds of FL.\\  
		\hline  
		$\theta _{n}^{t}$  & Local training model of user$_n$ in round \textit{t}.\\  
		\hline  
		$D _{\rm{aux}}$  & Auxiliary dataset held by the server.\\  
		\hline 
		$\alpha$  & Learning rate when server extracts sensitivity.\\  
		\hline 
		$m$  & Meta-classifier.\\  
		\hline
		$\rm{Agg()}$  & Model aggregation operations.\\  
		\hline
		$x$  & Number of models selected for aggregation.\\  
		\hline 
		$N_{\rm label}$  & Number of classes in the dataset.\\  
		\hline
		$S_{n}^{t}$  & Model sensitivity of user$_n$ at round \textit{t}.\\  
		\hline
		${\rm{DS}}_{n}^{t}$ & Differential model sensitivity of user$_n$ at round \textit{t}.\\
		\hline
		${\rm{Th}_{\rm round}}$ & Discrimination round threshold.\\  
		\hline
	\end{tabular}  
\end{table}  

\subsection{Meta-Classifier Offline Training} 

\begin{algorithm}
\small
	\caption{Training Meta-Classifier in Centralized Learning}
	\label{al2}
        \textbf{Input: Auxiliary dataset $D _{\rm{aux}}$, learning rate $\alpha$  } \\
	    \textbf{Output: Meta-classifier $m$} \\
		Sample the data in $D _{\rm{aux}}$ into multiple data distributions and train shadow models correspondingly \\
		Record $i_{\rm th}$ shadow model parameters ${\theta_{{\rm{shadow}}_i}}$ and its preference class ${\rm{mc}}_i$ \\
		Divide $D_{\rm{aux}}$ into $N_{\rm label}$ retraining datasets with all classes: $D_{\rm{aux}}\rightarrow \left \{ D _{\rm{aux}_{1}},..., D _{{\rm{aux}}_{t}},...,D_{\rm{aux}_{N_{\rm label}}} \right \}$ with $D _{{\rm{aux}}_{t}}$ having samples solely from the $t_{\rm th}$ class. \\
		Record retrained model parameters ${\theta_{{\rm{retrain}}_i}}$ and compute model sensitivity $S$ of its class $S = \sum\left | \left (\theta_{{\rm{retrain}}_i}- \theta_{{\rm{shadow}}_i}  \right )\cdot \frac{1}{{\alpha }} \right |$ \\
		Label the meta-data $\left(S , {{\rm{mc}}_i} \right)$, and train meta-classifier $m$ \\
\end{algorithm}

\begin{algorithm}
\small
	\caption{Training Meta-Classifier in Federated Learning}
	\label{al2.5}
        \textbf{Input: Auxiliary dataset $D _{\rm{aux}}$, learning rate $\alpha$  } \\
		\textbf{Output: Meta-classifier $m$} \\
		 Combine the data in $D_{\rm{aux}}$ into multiple data distributions and train shadow models respectively \\
		 Record preference class $\rm{mc}$ of each shadow model \\
		 Divide $D _{\rm{aux}}$ into $N_{\rm label}$ retraining datasets with all classes: $D_{\rm{aux}}\rightarrow \left \{ D _{\rm{aux}_{1}},..., D _{{\rm{aux}}_{t}},...,D_{\rm{aux}_{N_{\rm label}}} \right \}$ with $D _{{\rm{aux}}_{t}}$ having samples solely from the $t_{\rm th}$ class \\
    	 \textbf{Shadow Model Pairing:} \\
		      Aggregate each shadow model with its paired one that has opposite preference class, and record the aggregated model parameters for such shadow model ${\theta_{{\rm{agg}}_i}}$ \\
		      Retrain ${\theta_{{\rm{agg}}_i}}$ on per class subset $D _{{\rm{aux}}_{t}}$ to record model parameters ${\theta_{{\rm{agg}}_i}^{\rm{retrain}}}$ \\
		      Compute model sensitivity of this aggregated model $S1=\sum\left | \left ({\theta_{{\rm{agg}}_i}^{\rm{retrain}}}- {\theta_{{\rm{agg}}_i}}  \right )\cdot \frac{1}{{\alpha }} \right |$ \\
		\textbf{Shadow Model Updating:} \\
		     Update aggregated model ${\theta_{{\rm{agg}}_i}}$ on dataset corresponding to the $i_{\rm th}$ shadow model to resemble the local model training for the next round in FL, and record updated model parameters ${\theta_{{\rm{aux}}_i}}$ \\
		     Retrain ${\theta_{{\rm{agg}}_i}^{\rm{retrain}}}$ on per class subset ${\theta_{{\rm{aux}}_t}}$ to record model parameters ${\theta_{{\rm{aux}}_i}^{\rm{retrain}}}$ \\
		     Compute updated model sensitivity of its preference class $S2=\sum\left | \left ({\theta_{{\rm{aux}}_i}^{\rm{retrain}}}- {\theta_{{\rm{aux}}_i}}  \right )\cdot \frac{1}{{\alpha }} \right |$ \\
		     Label the meta-data $\left ( \left | S1-S2 \right |, \rm{mc} \right ) $, and train meta-classifier $m$ \\
\end{algorithm}

Before FL begins, the server needs to train a meta-classifier with the auxiliary dataset $D _{\rm{aux}}$, as detailed in Algorithm \ref{al2}. Firstly, the server samples the data in $D _{\rm{aux}}$ into a subset to resemble a heterogeneous data distribution of a user. The sampling repeats a number of times. For each subset, a shadow model is trained and its preferred class is acted as the label. Secondly, the server separates $D _{\rm{aux}}$ into $N_{\rm label} $ subsets $D_{\rm{aux}}\rightarrow \left \{ D _{\rm{aux}_{1}},..., D _{\rm{aux}_{t}},...,D_{\rm{aux}_{N_{\rm label}}} \right \}$ where $N_{\rm label} $ is the total number of classes, each subset consisting of samples from a single class. Then, each shadow model is retrained per subset ($D _{\rm{aux}_{t}}$) to extract shadow model sensitivity per class. Thirdly, such sensitivity (in particular, the gradient change) and its preferred class given a shadow model are used as the training set to train the meta-classifier. The meta-classifier can profile the preference class of the dataset. 

However, the meta-classifier trained in the above straightforward manner is only suitable for centralized deep learning not the collaborative learning, in particular, in FL. Because when a local model is aggregated, the data distribution characteristics will be concealed within the global model, resulting in the model sensitivity fading of each user preference class in the following round. To overcome this challenge, 
we improve Algorithm \ref{al2} to make the meta-classifier applicable to FL, as detailed in Algorithm \ref{al2.5} (the performance comparison between the baseline Algorithm \ref{al2} and Algorithm \ref{al2.5} are detailed in Section IV-B7). Generally, we leverage two specific improvements: paired aggregation, and differential sensitivity in two consecutive rounds. 

In the meta-classifier training of FL, each shadow model will be paired with another shadow model. The steps are as below.
\begin{enumerate}[noitemsep, topsep=2pt, partopsep=0pt,leftmargin=0.4cm]
    \item For the $i_{\rm th}$ shadow model with majority class ${\rm{mc}}_i$, the server selects the other shadow model with the highest model sensitivity of ${\rm{mc}}_i$ (i.e., ${\rm{mc}}_i$ is the minority class to a large extent) from all rest shadow models, and aggregates it with the $i_{\rm th}$ shadow model. 
    \item The server extracts the model sensitivity $S1$ of the aggregated model, and updates each aggregated model on the dataset corresponding to each shadow model. 
    \item Then the model sensitivity $S2$ of the updated $i_{\rm th}$ shadow model in the next round is extracted.
    \item The difference between $S1$ and $S2$ resembling two consecutive rounds in FL, $\left| S1-S2 \right|$, and its preferred class are meta-data used to train the meta-classifier in FL.
\end{enumerate}
Note in step 1), in the case of minority, the model with the lowest model sensitivity of ${\rm{mc}}_i$ model is selected for pairing. Also, the $S1$ is extracted from the aggregated model, while the $S2$ is merely extracted from the updated local model in the following round.

The meta-classifier will be used to accurately profile the user's preference during the online FL learning procedure when the sensitivity of user uploaded model is extracted and fed. The input of meta-classifier is the model sensitivity difference of two consecutive rounds ($\left| S1-S2 \right|$). In order to maximize the differential model sensitivity and make the meta-classifier get more accurate classification, we propose and leverage selective aggregation during the FL's training, which will significantly increase S1. Since S2 remains relatively unchanged regardless of whether the selective aggregation is employed, a higher differential model sensitivity will be generated in the next round to obtain accurate profiling.

\subsection{User Local Model Training}
FL is with a number of users in practice, and the users may be located in different regions, indicating statistical and system heterogeneity. The PPA-concerned statistical heterogeneity can be a result from personalized data that are collected~\cite{ZawadAC00BT021}.
Thus, each user's training set has its own preferred category (class). As shown in step (1) in Figure~\ref{scheme}, each user local data thus has its unique data distribution. Each user has a majority (minority) class ${\rm{mc}}_n$, that is, the largest (lowest) number of samples in the dataset. For example, label $1$ and label $6$ are the majority classes of user$_1$ and user$_n$, and label $4$ and label $7$ are the minority classes of user$_1$ and user$_n$, respectively, as exemplified in Figure~\ref{scheme}. 
In FL, the local model may be trained multiple epochs before it is uploaded to the server in an FL round. PPA has naturally tolerated such variation, which is invariant to the epoch number adopted by the user (see Section~\ref{sec:localepoch}). After local training, each user updates the model parameter $\theta _{n}^{t}$ and uploads it to the server at the $t_{\rm th}$ round. The local model inadvertently memorizes the preference of the corresponding user, which is the private information that the meta-classifier is interested in profiling later.

\subsection{Model Sensitivity Extraction}

To quantify the model sensitivity to preference class, we utilize sum of absolute values of the gradient changes before and after the local user model retraining, which is expressed:
\begin{equation}
    S= \sum\left | \frac{\delta }{\delta \theta _{n}^{t'}}L(\theta ) \right |=\sum\left | \left (\theta _{n}^{t}- \theta _{n}^{t'}  \right )\cdot \frac{1}{{\alpha }} \right |,
\end{equation}
where $\theta _{n}^{t}$ and $\theta _{n}^{t'}$ respectively represents the uploaded and retrained model parameter vector of user$_n$ at the $t_{\rm th}$ FL round, and \textit{L}($\cdot $) represents the loss function. Given a local model uploaded, the server retrains it based on the retrain dataset $\left \{ D _{\rm{aux}_{1}},..., D _{{\rm{aux}}_{t}},...,D_{\rm{aux}_{N_{\rm label}}} \right \}$ with $D _{{\rm{aux}}_{t}}$ to extract the sensitivity of the model per class. The $\theta _{i}^{\prime}$ denotes the retrained model. The process is detailed in Algorithm~\ref{alg:sensitivity}.

\subsection{Preference Class Profiling}
We launch PPA in the FL's training phase. In this phase, the server leverages a trained meta-classifier to profile user preference after extracting its uploaded model sensitivity. The meta-classifier profiles preference classes based on the differential model sensitivity between consecutive rounds. There are two considerations we take to improve the attack accuracy. 

Firstly, it may be inaccurate to determine the user's preference class merely according to sensitivity extracted from one FL round. Therefore, we determine the preference based on a number of consecutive rounds only if their preferences predicted by the meta-classifier are all same. This number of rounds is denoted as ${\rm{Th}_{\rm round}}$. Given meta-classifier outputs the same preference of a user in consecutively ${\rm{Th}_{\rm round}}$ FL rounds, this preference is locked, and PPA against this user is completed, which means the monitoring on this user is no longer needed. The process is detailed in Algorithm~\ref{alg:profilePreference}.

Secondly, the FL's aggregation can make inference attacks~\cite{Nasr2019Comprehensive,Melis2019Exploiting} challenging, as the global model merges all users' data characteristics. With one and more FL rounds, data characteristics of a given user are concealed and faded, thus preventing PPA from accurately extracting the user's model sensitivity that reflects the user's local data characteristics. To address this problem, we propose selective aggregation elaborated as follows.

\begin{algorithm}[t]
\small
    \caption{Extracting Model Sensitivity}	\label{alg:sensitivity}
		 \textbf{Input: Local models of all users in the round \textit{t} $\theta _{n}^{t}$, Auxiliary dataset $D_{\rm{aux}}$, learning rate $\alpha$ } \\
		 \textbf{Output: Model sensitivity of class $c$ $\textit{S}_{n}^{t}$} \\
		 Set  $\textit{S}_{n}^{t}=0$ \\
		 Generate retraining set of class $c$: $D_{{\rm{aux}}_{c}}$ \\
		 \ForEach{$\theta _{n}^{t}$ from 1 to $N_{\rm user}$ in the round \textit{t}} {
    	    Retrain model $\theta _{n}^{t'}= \theta _{n}^{t}. train(D_{{\rm{aux}}_{c}}, \alpha)$ \\
		    \ForEach {neuron $\textit{i}$ in $\theta _{n}^{t'}$} {
		        Compute $\textit{S}_{(n, i)}^{t}=\left | \left (\theta _{n}^{t}- \theta _{n}^{t'}  \right )\cdot \frac{1}{{\alpha }} \right |=\left | \frac{\delta }{\delta \theta _{i}}L(\theta ) \right |$   \\
		        $\textit{S}_{n}^{t} += \textit{S}_{(n, i)}^{t}$ \\
		    }
		 }
\end{algorithm}

\begin{algorithm}
\small
	\caption{Profiling Preference Class}	\label{alg:profilePreference}
 	 \textbf{Input: Model sensitivity $ \textit{S}_{n}^{t}$ of user$_n$ in the round \textit{t}, Model sensitivity $ \textit{S}^{t-1}$ of aggregated model in the last round \textit{t-1}, Round threshold ${\rm{Th}_{\rm round}}$} \\
		 \textbf{Output: Preference class of user$_n$ ${\rm{mc}}_n$} \\
		     Users train locally and upload models $\theta _{n}^{t}$ \\
		     Server retrains $\theta _{n}^{t}$ based on retraining dataset to get $\theta _{n}^{t'}$ \\
		     Extract model sensitivity $ \textit{S}_{n}^{t}$ of $\theta _{n}^{t}$ \\
		     Compute the differential model sensitivity between $ \textit{S}_{n}^{t}$ and $ \textit{S}^{t-1}$ resembling two consecutive rounds $\left |\textit{S}^{t-1}-\textit{S}_{n}^{t} \right |$ \\
		     Input differential model sensitivity into meta-classifier, and profile preference class ${\rm{mc}}_n$ \\
		     \If {same result in ${\rm{Th}_{\rm round}}$ rounds}{
		        PPA against this user is completed \\
		     }
		     Each uploaded model of user is aggregated with $x$ models that have the highest (lowest) model sensitivity of its majority (minority) class \\
		     Extract model sensitivity $\textit{S}^{t}$ of each aggregated model and then distribute such aggregated model back to corresponding user \\
\end{algorithm}

\subsection{Selective Aggregation}

There are two challenges of selective aggregation confronted by PPA: 1) how to prevent the user model sensitivity from being decreased or concealed by the global aggregation process after a number of FL rounds? 2) how to further amplify the user model sensitivity to reflect the user's specific local data characteristics?

\begin{figure}
	\centering  
	\includegraphics [width = 0.33\textwidth]{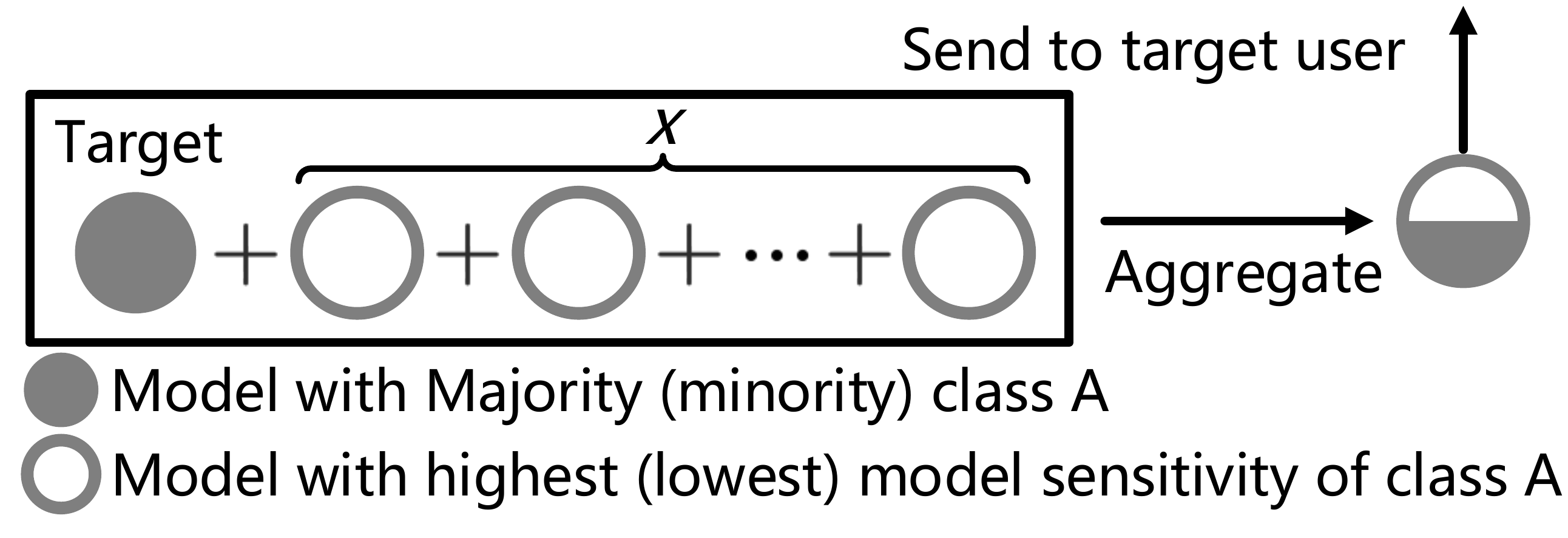}  
	\caption{Selective aggregation. The selected $x$ users have \textit{opposite model sensitivity} of class A that is the majority (minority) class of the targeted user.}  
	\label{Selective}  
\end{figure}

In the selective aggregation, given a targeted user, the server selects models of $x$ users who exhibit an opposite majority (minority) class as their models have the opposite highest (lowest) extracted model sensitivity---note lowest sensitivity corresponds to majority class. 
Then these $x+1$ models are aggregated to form a global model, which is then sent back to the targeted user only, as shown in Figure~\ref{Selective}---for other users who are not interested by the attacker, the global model sent to them follows a normal aggregation. After selective aggregation, the server extracts the model sensitivity of each aggregated model for the next round of meta-classifier profiling.
The selective aggregation is inconspicuous to the local users due to two main facts. Firstly, users only communicate with the server rather than other users, and the communication between a user and the server is usually through a standard secure communication channel. This is important to prevent privacy leakage (e.g., property inference attack~\cite{Melis2019Exploiting}) due to a victim user's local model direct exposure to a malicious user---attack occurred between users.
Secondly, FL can indeed distribute different global models to users, such as personalized FL~\cite{PersonalizedFL} that leverages a clustering technique to group users and sends different global models to different groups of users.
Therefore, it is tractable to make the users unaware of selective aggregation. Later we experimentally show that the local data accuracy predicted by the global model (i.e., model utility) is similar from the user's perspective with and without implementing the selective aggregation.

The input of meta-classifier is the model sensitivity difference of two consecutive rounds. We define it as differential model sensitivity ($DS$), and compute as follows:
\begin{equation}
DS_{n}^{t} =\left | S_{{\rm{agg}}_{n} }^{t-1}-S_{n}^{t}\right |,
\end{equation}
where $S_{{\rm{agg}}_{n} }^{t-1}$ is the model sensitivity of the \textit{aggregated model} in round $t-1$, and $S_{n}^{t}$ is the model sensitivity of the updated user model in round $t$. Through selective aggregation, $S_{{\rm{agg}}_{n} }^{t-1}$ of each model will be amplified, while the $S_{n}^{t}$ is stable compared to the case without the usage of selective aggregation. Therefore, the $DS$ in each round is greatly improved since the meta-classifier is trained based on $DS$, the larger the $DS$, higher the profiling accuracy. Notably, in order to maximize the $DS$ in the next round of PPA, the server utilizes selective aggregation by sending different models to each user. 

\section{Experiments}\label{sec:experiment}
In this part, we experimentally evaluate PPA and analyze its attack performance under four data heterogeneity metrics (\textit{CP}, \textit{CD}, \textit{UD}, and \textit{ID}) in FL. In addition, we validate the effectiveness of selective aggregation under a number of varying settings. Moreover, the scalability of PPA is examined and validated.

\subsection{Experimental Setup}

In our experiments, common personal computers act as users in FL with an identical configuration (as system heterogeneity is not our concern): Intel Core i5 processors for a total of four cores running at 3.20 GHz and 12\,GB RAM. We use H3C UIS 3010 G3 8SFF server as the FL's aggregator (namely the attacker), which is equipped with two Intel Xeon CPU Silver 4214, 128\,GB RAM and two 480\,GB SSD. 

To evaluate the efficacy of PPA inference, we conduct experiments on four different datasets. In particular, the Products-10K is used to represent a commercial scenario and the facial expression dataset (i.e., RAF-DB) represents personal private information.

\subsubsection{\textbf{Datasets and Model Architecture}}
We select four datasets to evaluate PPA. MNIST and CIFAR10 are used to comprehensively show the PPA's performance under four data heterogeneity metrics and the effectiveness of selective aggregation. Products-10K and RAF-DB are real-world datasets used to indicate the PPA's performance under commercial scenarios of shopping and social network and the scalability of PPA.

\begin{itemize}[noitemsep, topsep=2pt, partopsep=0pt,leftmargin=0.4cm]
	\item \textit{MNIST} is an image dataset of handwritten numbers from 0 to 9~\cite{MNIST}.	The images in the database are all 28$\times$28 grayscale images, including 60,000 training images and 10,000 test images.
	
	\item \textit{CIFAR10} is an image dataset for the recognition of universal objects~\cite{CIFAR10}. There are 10 categories of RGB color images	with size of is $32 \times 32 \times 3$. It consists of 50,000 training and 10,000 testing samples, respectively.
	
	\item \textit{Products-10K} is a product identification set built by JD AI research, which contains about 10000 products often purchased by Chinese consumers and covers many categories, e.g., fashion, food, health care, household products, etc~\cite{2020Products}. We use this dataset to resemble a real-world FL shopping scenario and evaluate the impact of our attack.
	\item \textit{RAF-DB}, Real-world Affective Faces Database~\cite{li2017reliable}, is a large-scale facial expression database with around 30K great-diverse real-world facial images. Images in this dataset are of the great variability in subjects' age, gender and ethnicity, head poses, lighting conditions, occlusions, post-processing operations, etc. 
\end{itemize}

The model architecture for training MNIST is a convolutional neural network with two convolution layers, followed by a max pooling layer, and then a fully connected layer. As for the remaining CIFAR10, Products-10K, and RAF-DB, the model architecture has four convolution blocks: each block has one convolutional layer and one max pooling layer, followed by two fully connected layers. We note that the model architectures are well aligned with previous orthogonal privacy inference attacks. Particularly, the most complicated model architecture that previous orthogonal privacy inference attacks~\cite{Ganju2018Property, Hitaj2017GAN, Melis2019Exploiting, Shokri2017member} used  is a CNN with three convolutional layers. PPA is applicable to complicated model architectures (e.g., VGG), as validated in Section~\ref{sec:complicated}.

\subsubsection{\textbf{Data Distribution Metrics}}
Four metrics defined below are used to resemble the statistical heterogeneity of local user data in FL. Generally, the first two metrics are utilized to evaluate the PPA's performance under a varying data distribution given a user, and the last two are used to validate the applicability of the selective aggregation. Note that exhaustively experimenting all combinations of four data distributions is infeasible, we thus consider typical (common) combinations of them (the range of each metric's distribution is carefully considered e.g., each \textit{CP}, \textit{CD}, \textit{UD} from 10\% to 100\%).

\begin{itemize}[noitemsep, topsep=2pt, partopsep=0pt,leftmargin=0.4cm]
	
	\item \textbf{Class Proportion (\textit{CP})} represents the proportion of each class in the user dataset. Closer the value to 1, more samples in this class. The \textit{CP} is expressed as: 
	\begin{equation}
        CP =\frac{\#  \textit{number of samples given a class}}{\# \textit{dataset size}} 
	\end{equation}
	\item \textbf{Class Dominance (\textit{CD})} represents the dominance of a class in the user dataset. The numerator is the difference between the majority (minority) number and the class closest to it, expressed as below: 
	\begin{equation}
	    CD =\frac{\# \, |Majority (Minority) \,\, -\,\,Secondary|}{\# \, dataset \,\, size} 
	\end{equation}
	\item \textbf{User Dispersion (\textit{UD})} represents the divergence of preferred majority classes among users in FL. Numerator is the maximum difference in the number of users with the same preference class. Take an example to ease the understanding, there are three classes: A, B, C. Out of a total of 10 users there are 5 users with A as the preference class, 3 with B and 2 with C. The \textit{UD} = $(5-2)/10 = 30\%$. The numerator in this case is $5-2=3$ and the denominator is $10$. The closer the \textit{UD} to 1, the more consistent the user preferences in FL, expressed as below: 
	\begin{equation}
	    UD =\frac{\# \, range(user \,\, number)}{\# \, user \,\, number} 
	\end{equation}
	\item \textbf{Imbalance Degree (\textit{ID})} represents the degree of user data heterogeneity in FL, and measures the variance of local user \textit{total} dataset volume (size). More specifically, different users have different number of local samples, e.g., user A having 1,000 and user B having a differing 2,000. A larger value means a greater variance of user data volume and a higher degree of data heterogeneity, computed as follows: 
	\begin{equation}
	    ID={\# \, s^2 \,\, (user\,\,dataset\,\,volume)} 
	\end{equation}
\end{itemize}

\begin{figure*}
	\centering
	\subfigure[Class Proportion]{
		\includegraphics[width = 0.23\textwidth]{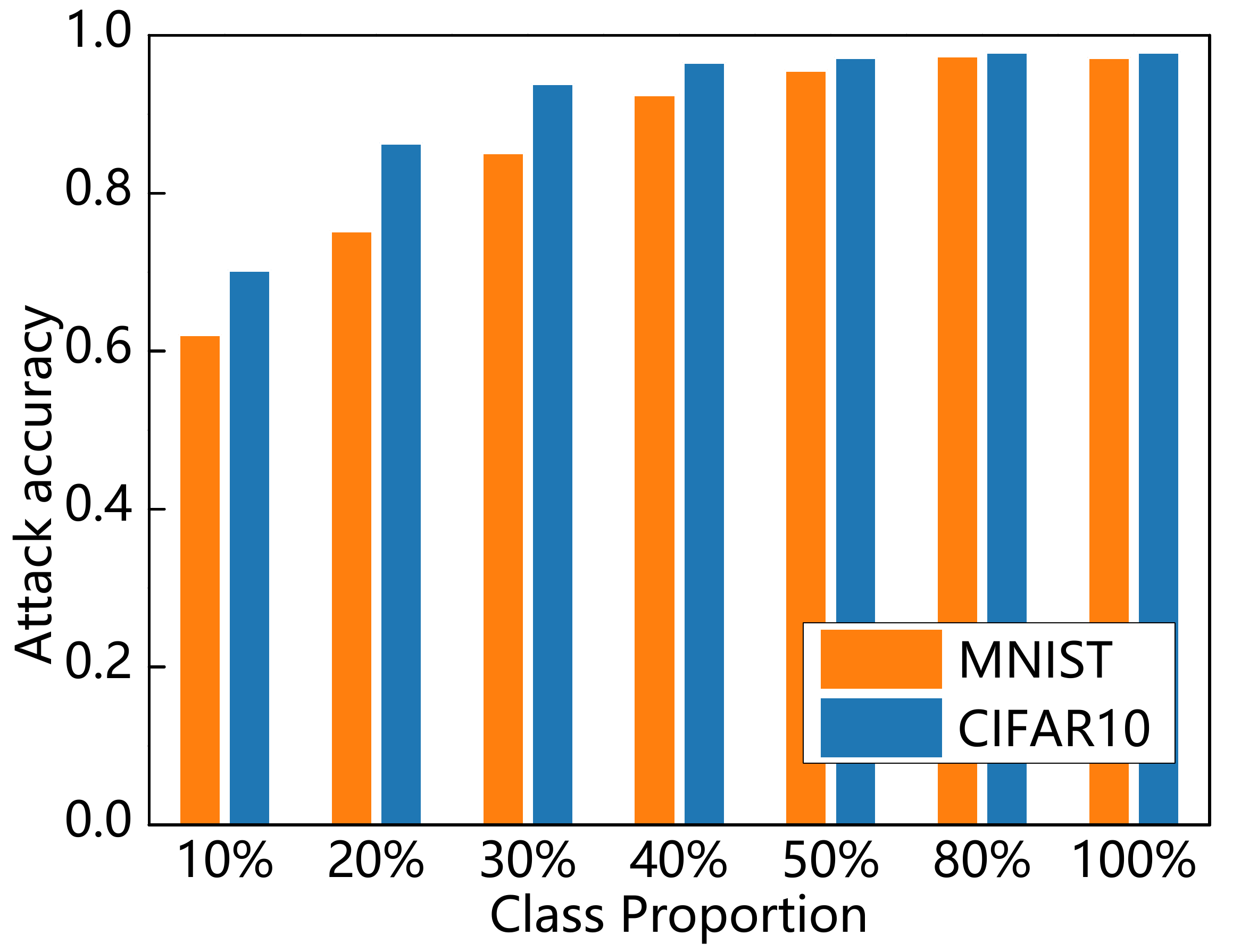}
		
	}
	\subfigure[Class Dominance]{
		\includegraphics[width = 0.233\textwidth]{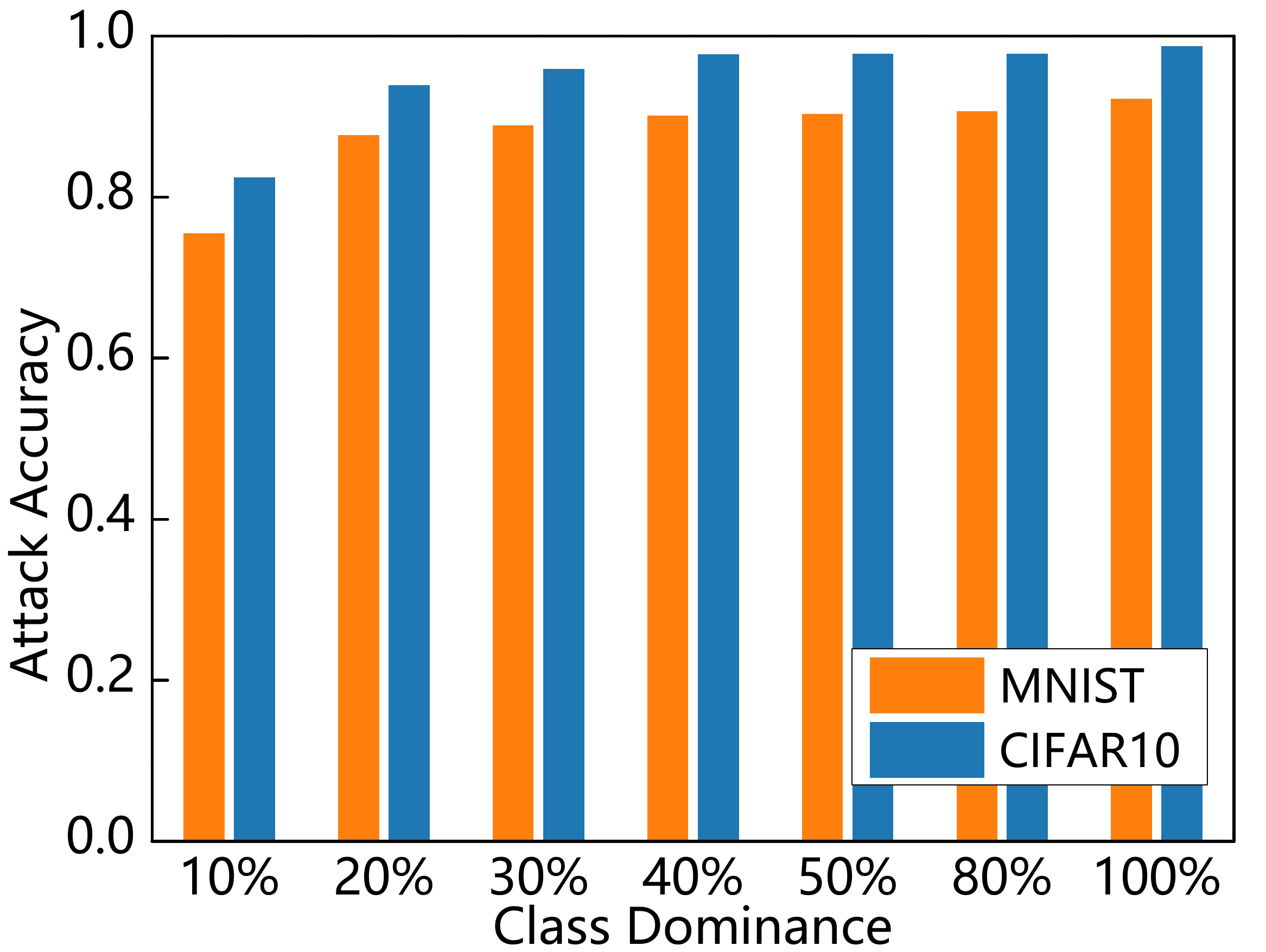}
	}
	\subfigure[User Dispersion]{
		\includegraphics[width = 0.23\textwidth]{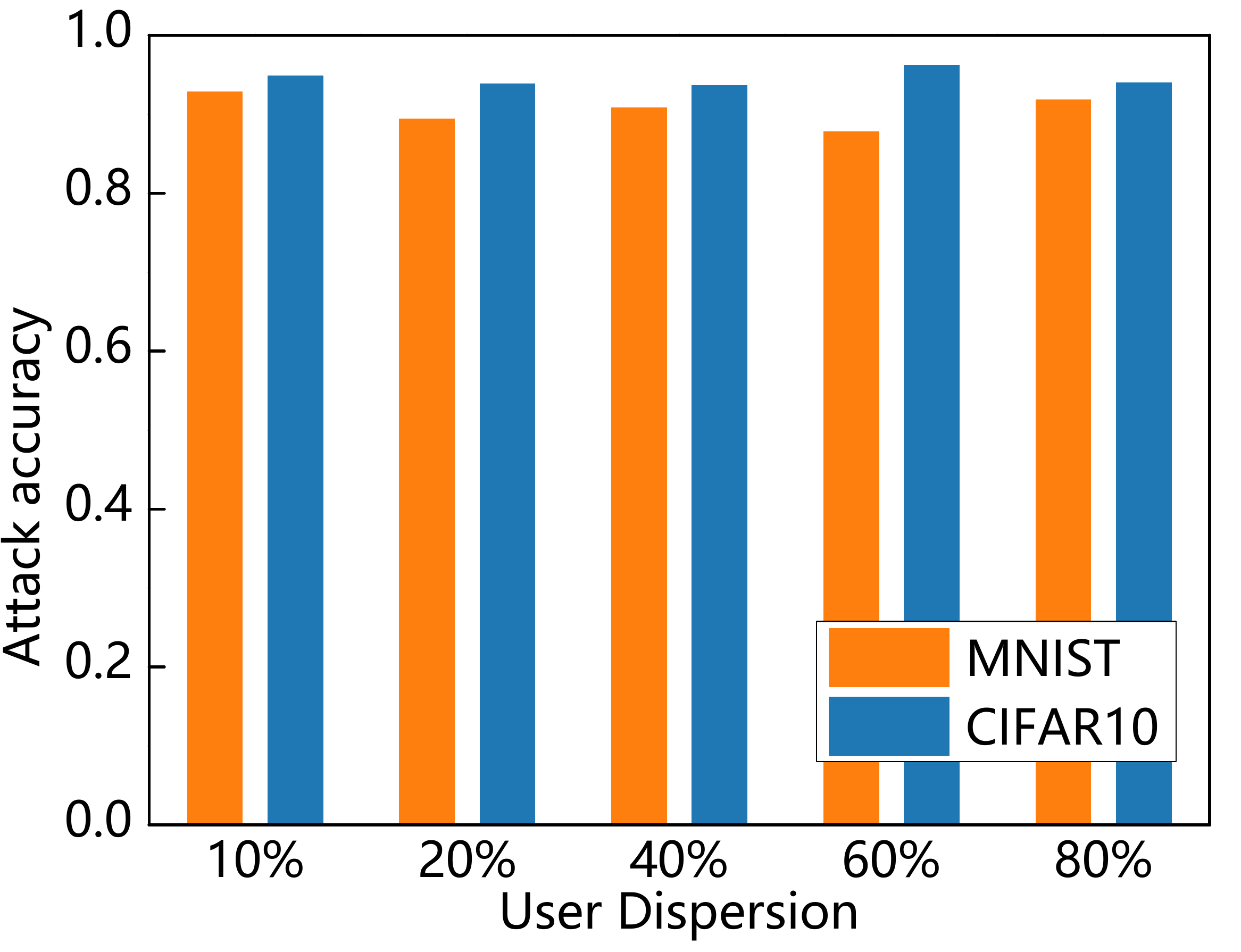}
	}
	\subfigure[Imbalance Degree]{
		\includegraphics[width = 0.23\textwidth]{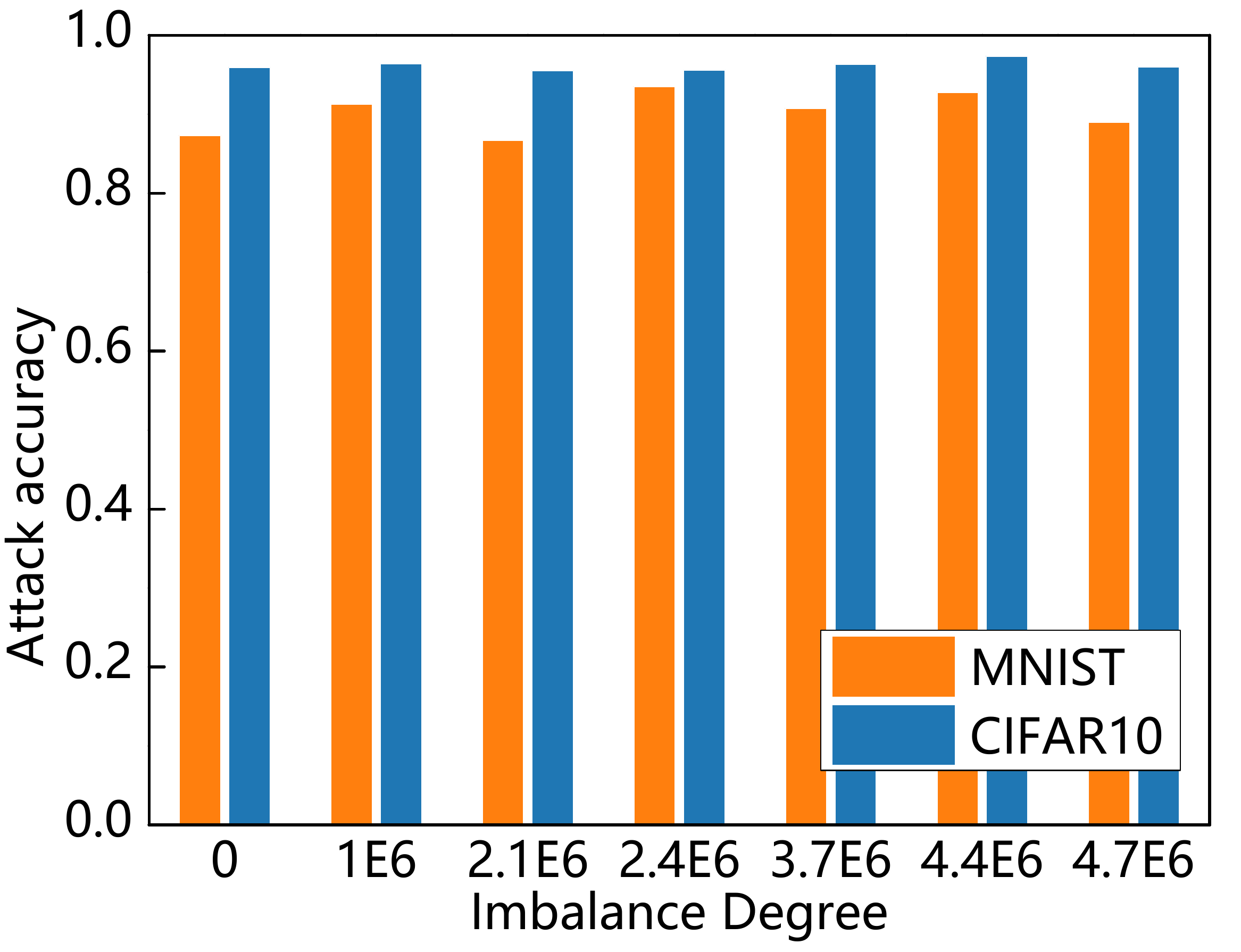}
	}
	\caption{Attack accuracy of PPA under different metrics. }
	\label{attackacc}
\end{figure*}

\subsection{Attack Performance}
Here, we use CIFAR10 and MNIST datasets for a extensive evaluation of the PPA's performance  under each of the aforementioned metrics. PPA is mainly evaluated against majority class. If the majority class predicted by meta-classifier is not the true majority class of the targeted user, \textit{even this class is predicted as the second majority class}, this PPA attempt is regarded as a failure. It should be noted that in practice, inferring the second majority class is still valuable for the attacker, e.g., the most and the second most popular items of a shopping mall. This top-$k$ profiling accuracy with $k$ no less than 1 is evaluated in Section~\ref{sec:realworld}.

For all experiments unless otherwise specified, we set up a total number of 10 users in FL. Each local dataset has 4,000 samples. We conduct multiple evaluations under the four metrics to entirely include the general data distribution in FL \cite{McMahan2017Communication,Melis2019Exploiting,Nasr2019Comprehensive}. As for the server, it has 150 auxiliary samples per class that are non-overlapped with any user. During PPA, the server applies selective aggregation (we compare it with normal aggregation in Section~\ref{sec:selectiveComp}).

\subsubsection{\textbf{Attack Accuracy under \textit{CP}}}
We decrease the \textit{CP} from 100\% to as low as 10\%. A higher \textit{CP} means that the preferred class has a larger number of samples, thus increasing the model sensitivity gap between this majority class and the rest classes. Therefore, the attack accuracy would be improved because of easier differentiating the majority class by the meta-classifier. The results of MNIST and CIFAR10 are displayed in Figure~\ref{attackacc} (a), which do confirm this tendency. Meanwhile, CIFAR10 performs better than MNIST because the variation in model sensitivity becomes larger as the dataset becomes richer in features. Each cases' accuracy is beyond 90\% when the \textit{CP} is higher than 40\%. 

\begin{figure*}
	\centering
	\subfigure[MNIST]{
		\includegraphics[width = 0.23\textwidth]{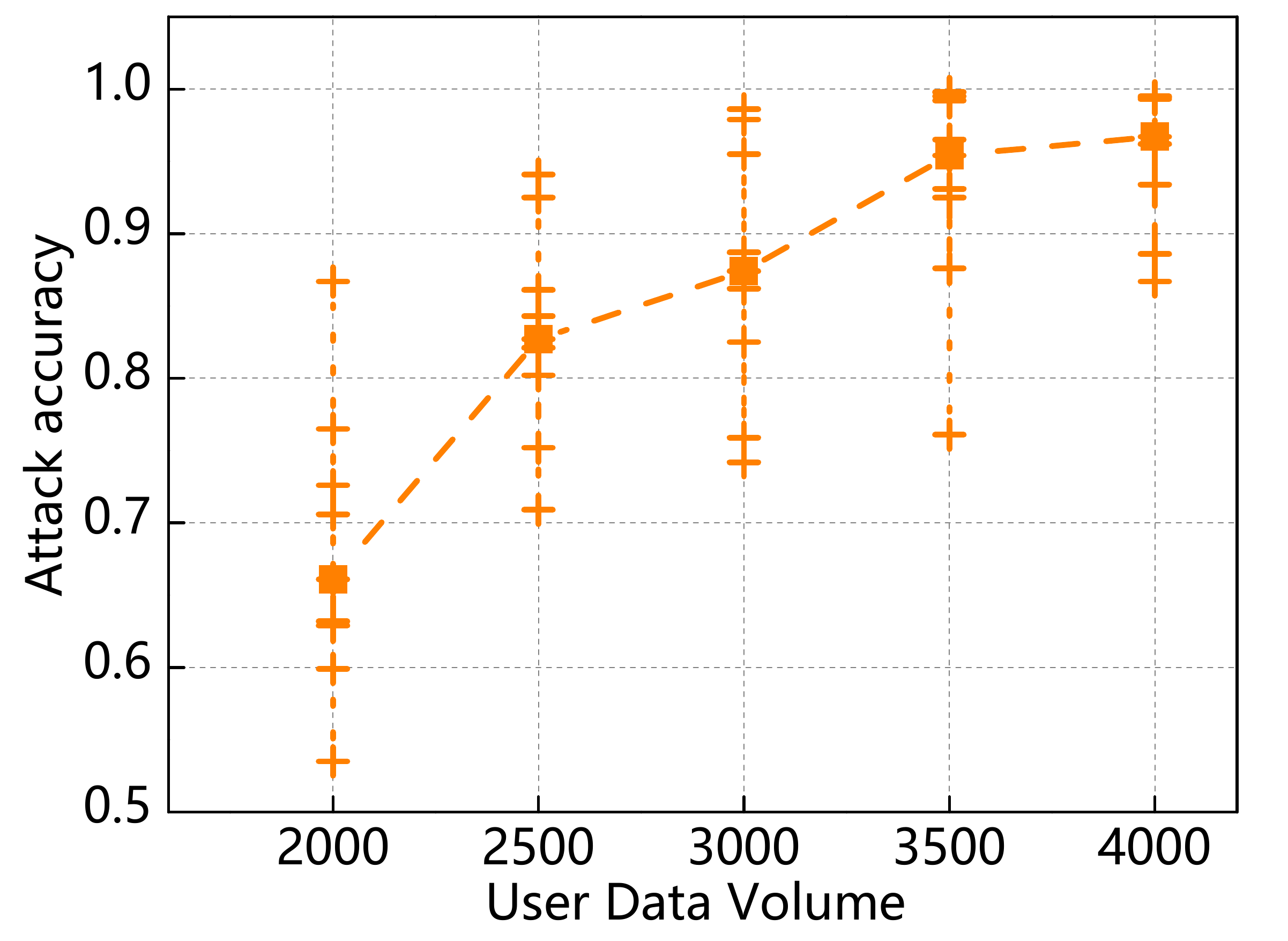}
		
	}
	\subfigure[CIFAR10]{
		\includegraphics[width = 0.23\textwidth]{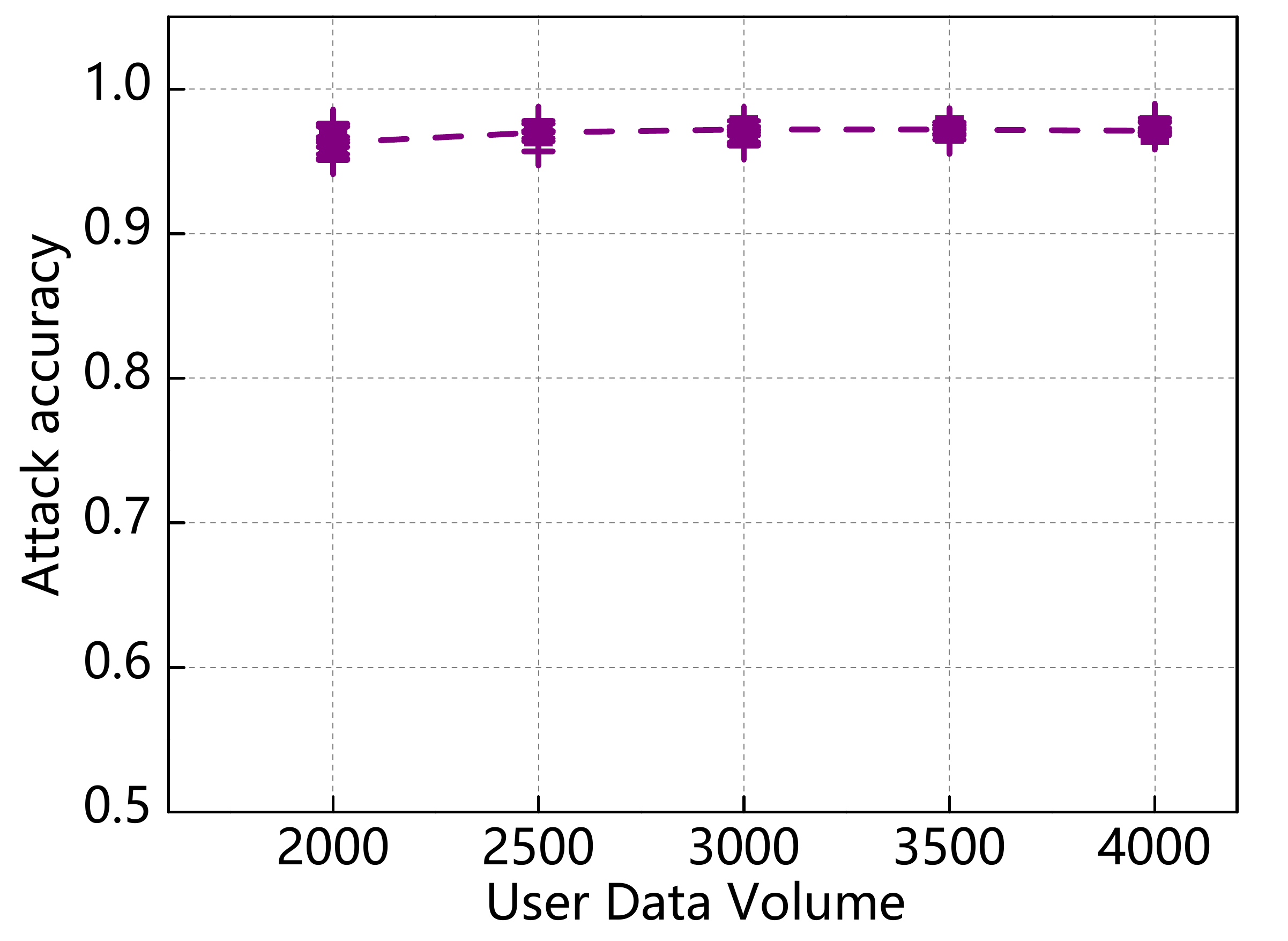}
	}
	\subfigure[MNIST]{
		\includegraphics[width = 0.23\textwidth]{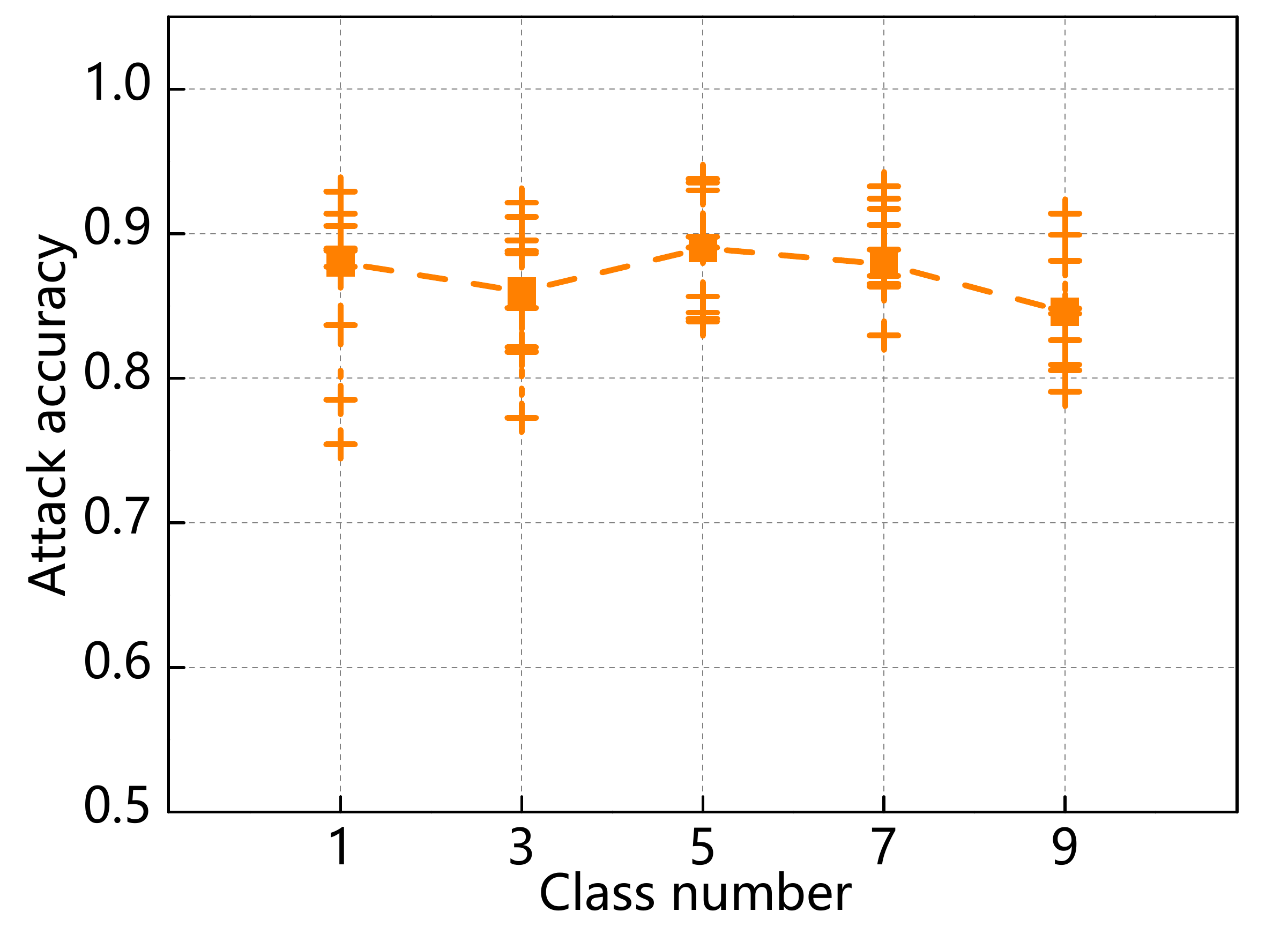}
	}
	\subfigure[CIFAR10]{
		\includegraphics[width = 0.23\textwidth]{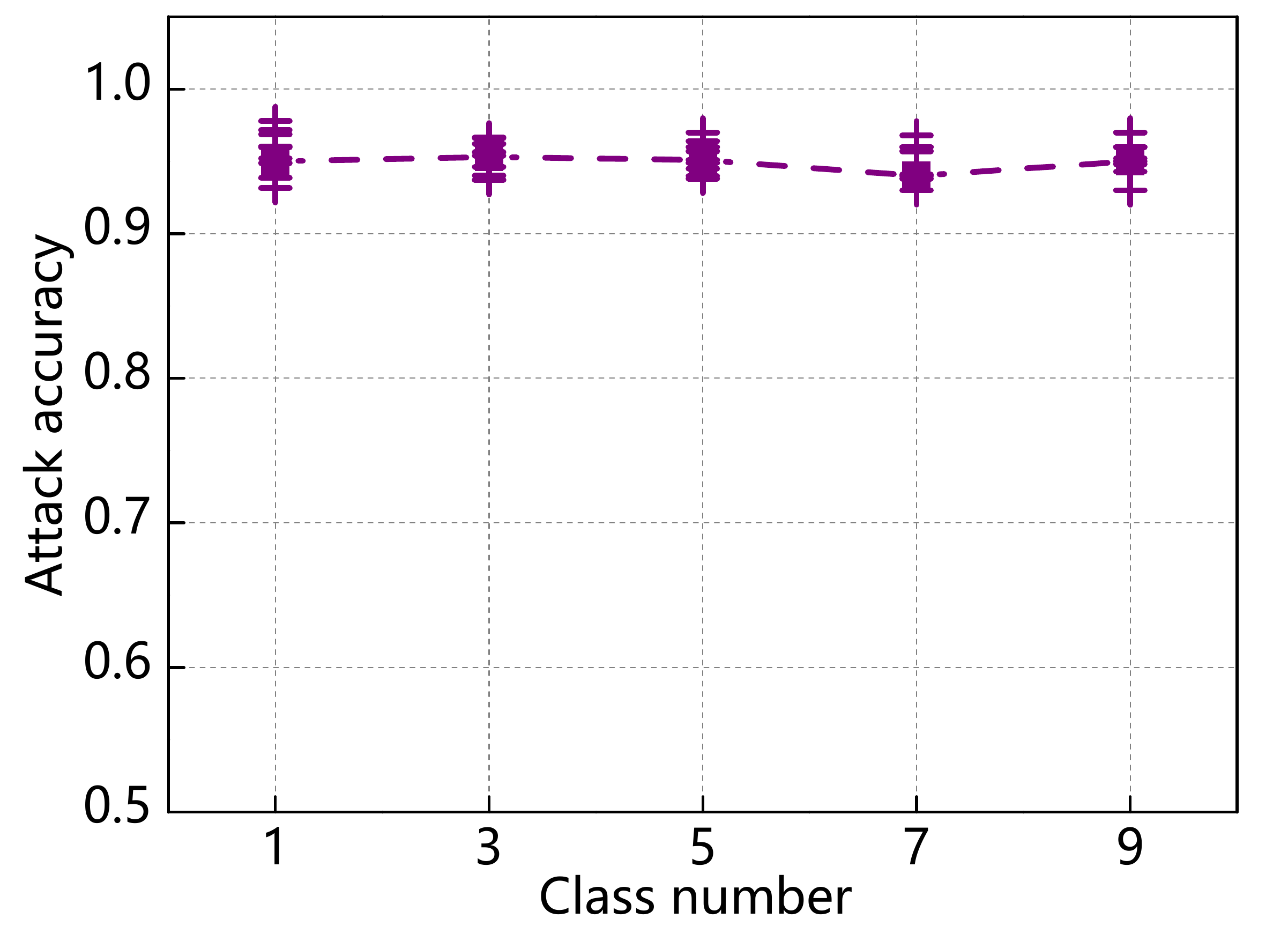}
	}
	\caption{Attack accuracy of PPA against models trained on MNIST and CIFAR10 datasets. Figure (a) and (b) reveal the attack accuracy with different amounts of data. The median values are connected across different training set sizes. Figure (c) and (d) reveal the accuracy of the attacks with different number of classes.}
	\label{mid}
\end{figure*}

\subsubsection{\textbf{Attack Accuracy under \textit{CD}}}
Here we vary the \textit{CD} from 10\% to 100\% and PPA's results for MNIST and CIFAR10 are detailed in Figure~\ref{attackacc} (b). When \textit{CD} is close to 0\%---the number of majority class is close to that of second majority class, the PPA's accuracy for the majority class is low. This is because the model sensitivity extracted for majority and second majority classes are similar, making the meta-classifier to be confused to determine the truth one. Nonetheless, as \textit{CD} increases, the attack accuracy greatly improves.

From the experimental results under \textit{CP} and \textit{CD}, we can see that PPA effectively profiles a user's preference within a wide range of data heterogeneity. In addition, richer characteristics of the data (i.e., CIFAR10) indicate a more effective meta-classifier for performing PPA.

\subsubsection{\textbf{Attack Accuracy under \textit{UD}}}
\textit{UD} indicates the degree of variation in the preferences of different user groups, reflecting the overall trend of preference categories in FL. It may affect the selection of user models in each round of selective aggregation, thus we evaluate the relationship between different \textit{UD} and attack accuracy, as shown in Figure~\ref{attackacc} (c). Whether the group preference is more consistent (\textit{UD} close to 1) or  more discrete (\textit{UD} close to 0), PPA can accurately profile the preference class of each user in FL.

\subsubsection{\textbf{Attack Accuracy under \textit{ID}}}
To affirm that selective aggregation is suitable for FL under imbalanced data, we evaluate the performance of PPA as a function of varying \textit{ID}, as shown in Figure~\ref{attackacc} (d). The experimental results indicate that the data amount owned by a user does not affect the effectiveness of PPA. 

Through the evaluations under \textit{UD} and \textit{ID}, the scalability of selective aggregation leveraged by PPA is validated.

\begin{table}
\small
	\begin{center}
		\caption{PPA attack accuracy vs model utility.} 
		\label{tab:acc1}
		\begin{tabular}{c|c|c|c|c}
			\hline
			& \multicolumn{2}{c|}{\centering Attack Accuracy}  &\multicolumn{2}{c}{\centering Model Utility}\\
			\hline 
			{Dataset} &\makecell[c]{ Train } 
			& \makecell[c]{ Test }
			& \makecell[c]{ without Attack} &  \makecell[c]{with Attack}  \\
			
			\hline  
			{MNIST}   & 0.994 & 0.894 & 0.913 & 0.902 \\  
			
			{CIFAR10}   & 0.981 & 0.967 & 0.821& 0.813 \\
			\hline  
			
		\end{tabular}
	\end{center}
\end{table}

\begin{figure}
	\centering
	\includegraphics[width = 0.3\textwidth]{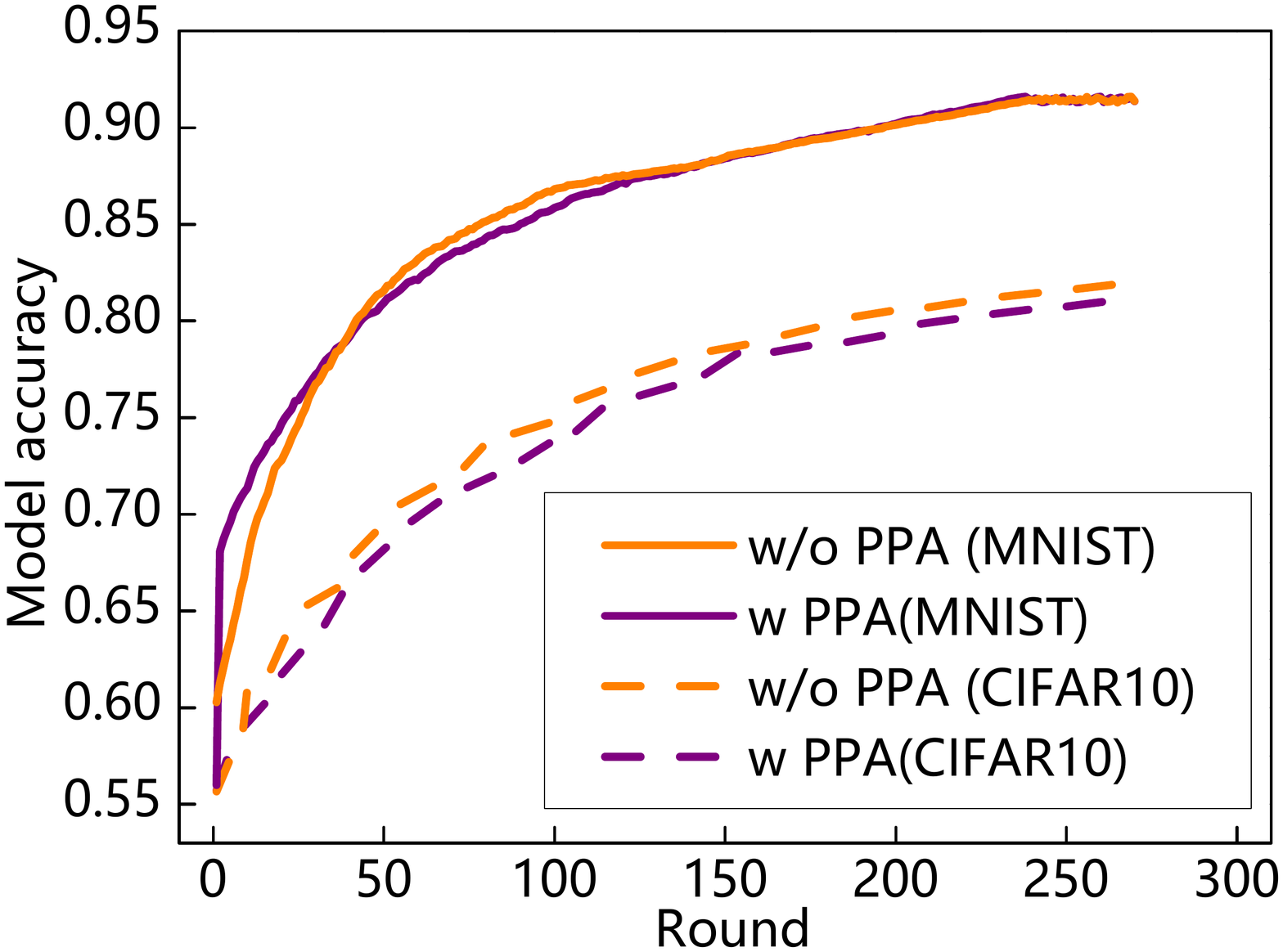}
	\caption{A comparison of model utility w/o the attack.}
	\label{useracc}
\end{figure}

\subsubsection{\textbf{Local Data Size and Class Number}}
In previous experiments, the local data size and number of classes in the local dataset are constant to be 4,000 and 10, respectively. We here vary these two parameters (note \textit{CP} is retained as 40\% and \textit{CD} is retained as 50\%). Specifically, for user data volume, we vary it from 2,000 to 4,000 by a step of 500. For the number of classes owned by a user, we vary it from 1 to 9 with a step by 2. We repeat 10 times for each setting. The attack accuracy is shown in Figure~\ref{mid}. The results indicate that the PPA's performance is independent of the number of classes of local data. As for the local data size, it to some extent influences the PPA's accuracy, especially when the sample itself (i.e., MNIST) is with a less rich feature. However, the local data size has a very limited impact on the attack accuracy given that the CIFAR10 is with richer features.

\subsubsection{\textbf{Attack Accuracy vs Model Utility}}
To be stealthier, the global model utility downloaded by the user should not be affected by  PPA. Otherwise, the user may be aware of the abnormality during the FL's training such as deviated model accuracy (utility) resulted from the selective aggregation. 
In this context, we evaluate the user model utility (i.e., its accuracy) in each round. In addition, we report the attack performance.

As summarised in Table \ref{tab:acc1} (\textit{CP} is 40\% and \textit{CD} is 50\%), for attacking performance, we evaluate the training and testing accuracy of meta-classifier. For model utility, we compare the accuracy of the user's local model before uploading in each round with the accuracy of the received aggregation model (after attack) in the following round. The experimental results show that given the often heterogeneous data distribution, PPA has the attack accuracy of 0.894 and 0.967 in MNIST and CIFAR10. Most importantly, there is no notable model utility difference without and with attack (specifically, the applied selective aggregation), which obviates the user's awareness of PPA by solely examining the model utility.
We also compare the change of user model accuracy with the number of rounds with and without the attack, as shown in Figure~\ref{useracc}. The experimental results reveal that there is no difference in user model accuracy between the two cases because PPA does not tamper with the user model parameters. Therefore, the model with PPA does not differ notably in accuracy from normal FL that may raise suspicion among users, which satisfies attack stealthiness.

\subsubsection{\textbf{Attack Performance: Algorithm 1 vs Algorithm 2}}
To validate the significant performance of PPA improved from Algorithm 1 to Algorithm 2, we evaluate the loss and accuracy of meta-classifier per algorithm.  The results are detailed in the Table \ref{baseline}, which affirm that meta-classifier in Algorithm 2 exhibits greatly improved performance over Algorithm 1: lower loss and higher accuracy. Because in FL scenario, Algorithm 2 utilizes selective aggregation to extract \textit{DS}, which is used to train meta-classifier. It is more suitable for distinguishing users with different preference classes in multiple users scenarios.


\begin{table}[]
\centering
\caption{A comparison of the PPA's performance between Algorithm 1 and Algorithm 2 in terms of loss and accuracy.
} 
\label{baseline}
\begin{tabular}{c|cc|cc}
\hline
\multicolumn{1}{l|}{} & \multicolumn{2}{c|}{Loss}                                         & \multicolumn{2}{c}{Accuracy}                                     \\ \hline
\multicolumn{1}{l|}{} & \multicolumn{1}{c|}{Algorithm 1} & \multicolumn{1}{l|}{Algorithm 2} & \multicolumn{1}{l|}{Algorithm 1} & \multicolumn{1}{l}{Algorithm 2} \\ \hline
MNIST                 & \multicolumn{1}{c|}{2.23}       & 1.34                            & \multicolumn{1}{c|}{0.43}       & 0.85                           \\ \hline
CIFAR10               & \multicolumn{1}{c|}{0.81}       & 0.49                            & \multicolumn{1}{c|}{0.71}       & 0.82                           \\ \hline
Products-10K           & \multicolumn{1}{c|}{1.89}       & 0.77                            & \multicolumn{1}{c|}{0.51}       & 0.84                           \\ \hline
RAF-DB                & \multicolumn{1}{c|}{0.88}       & 0.64                            & \multicolumn{1}{c|}{0.66}       & 0.81                           \\ \hline
\end{tabular}
\end{table}

\begin{figure*}
	\subfigure[Model sensitivity change under two aggregations (MNIST)] 
	{
		\begin{minipage}{8cm}
			\centering         
			\includegraphics[scale=0.3]{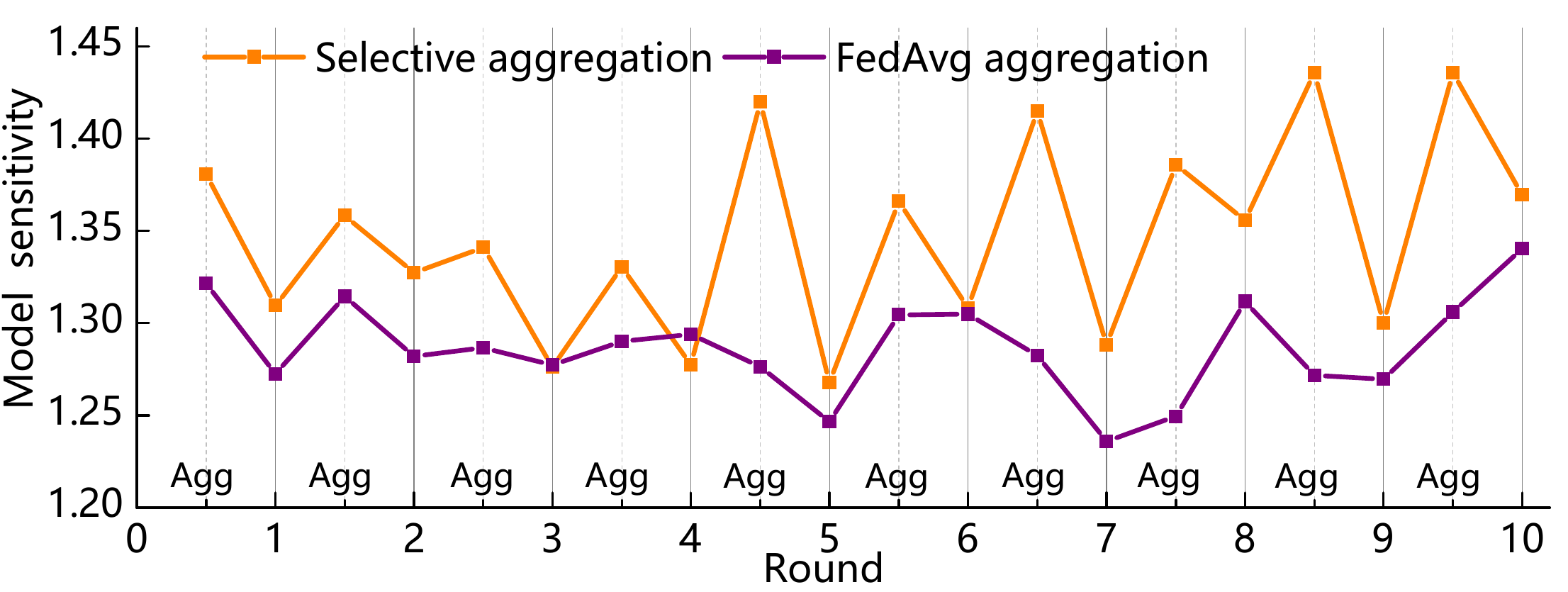}   
		\end{minipage}
	}
	\subfigure[Model sensitivity change under two aggregations (CIFAR10)] 
	{
		\begin{minipage}{8cm}
			\centering      
			\includegraphics[scale=0.3]{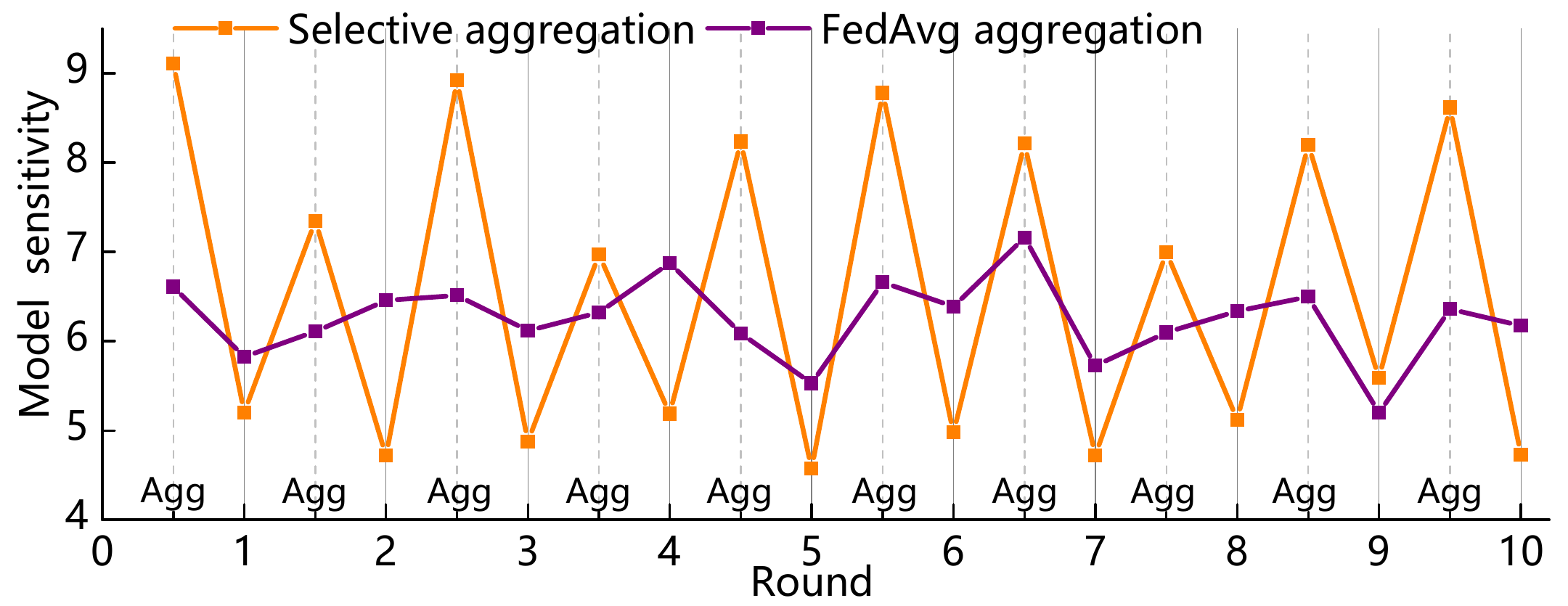}   
		\end{minipage}
	}
	\subfigure[Differential model sensitivity under two aggregations (MNIST)] 
	{
		\begin{minipage}{8cm}
			\centering          
			\includegraphics[scale=0.3]{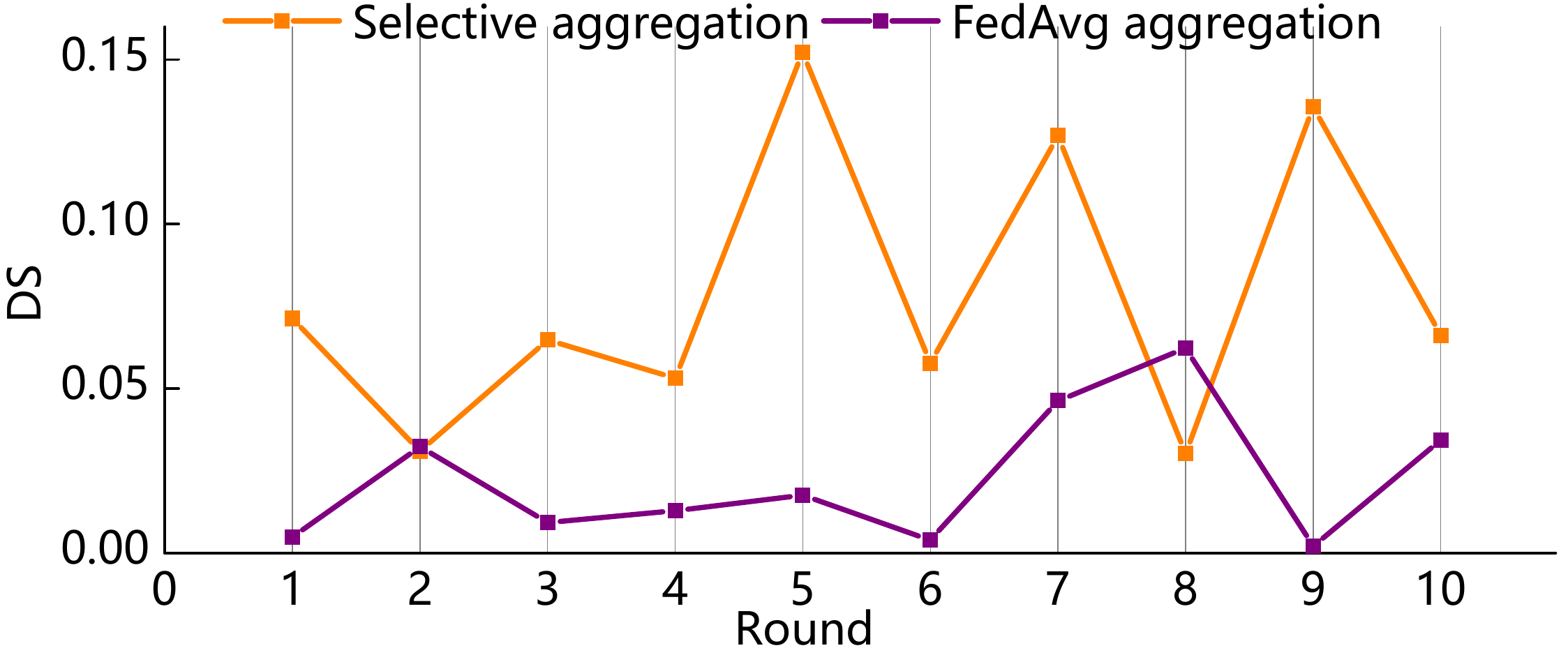}   
		\end{minipage}
	}
	\subfigure[Differential model sensitivity under two aggregations (CIFAR10)] 
	{
		\begin{minipage}{12cm}
			\centering      
			\includegraphics[scale=0.3]{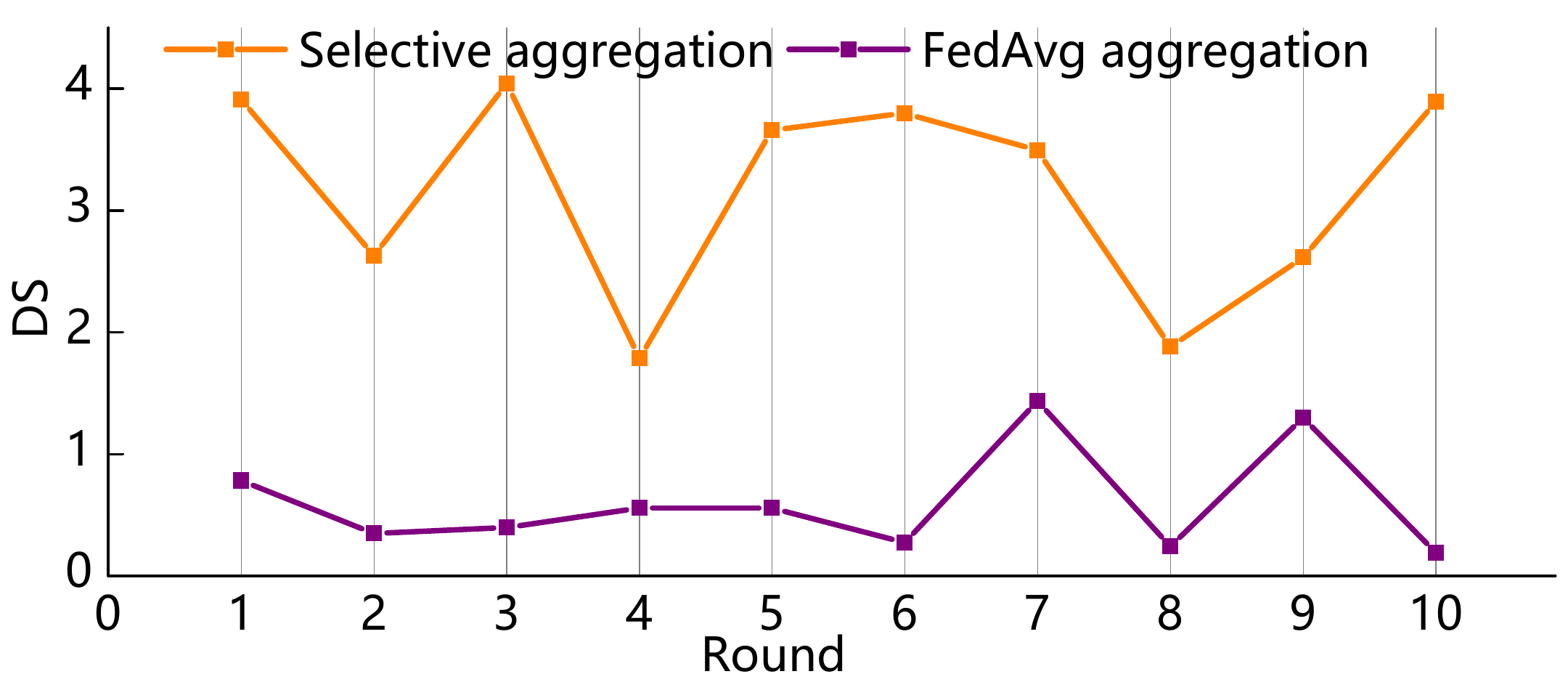}   
		\end{minipage}
	}
	\caption{Model sensitivity change and differential model sensitivity under two aggregations: with and without selective aggregation.} 
	\label{wave}  
	
\end{figure*}

\subsection{w/ and w/o Selective Aggregation} \label{sec:selectiveComp}

So far, all experiments are with the selective aggregation. We have not validated the improved efficiency of the selective aggregation compared to the case where the selective aggregation is not enforced and instead a common aggregation algorithm (i.e., \texttt{FedAvg}) is used. In the following, we compare the PPA's performance with and without selective aggregation from four aspects.

\subsubsection{\textbf{Differential Model Sensitivity}}

Each round of FL consists of two steps: model aggregation by server and local model updating by user. In each step per round, we extract the model sensitivity of the majority class under selective aggregation and \texttt{FedAvg} aggregation respectively, as shown in Figure~\ref{wave}(a) and (b). Then, we also compare the differential model sensitivity under two aggregations, as shown in Figure~\ref{wave}(c) and (d). 

As can be seen from Figure~\ref{wave} (a) and (b), each round of selective aggregation increases the model sensitivity. Then the next round of local training draws down the model sensitivity again. Though the model sensitivity of the \textit{user uploaded model},  $S_{agg_{n}}$, under \texttt{FedAvg} aggregation has no notable difference from the selective aggregation method. The model sensitivity of the aggregated model, $S_{agg_{n} }^{t-1}$, does have a larger difference. Since $DS_{n}^{t} =\left | S_{agg_{n} }^{t-1}-S_{n}^{t}\right |$, the $DS_{n}^{t}$ of selective aggregation method is greatly amplified compared to that of the \texttt{FedAvg} aggregation method, which can be clearly observed from Figure~\ref{wave}(c) and (d) that the $DS_{n}^{t}$ of selective aggregation method is always above that of the \texttt{FedAvg}.

\begin{figure}
	\centering  
	\includegraphics [width = 0.3\textwidth]{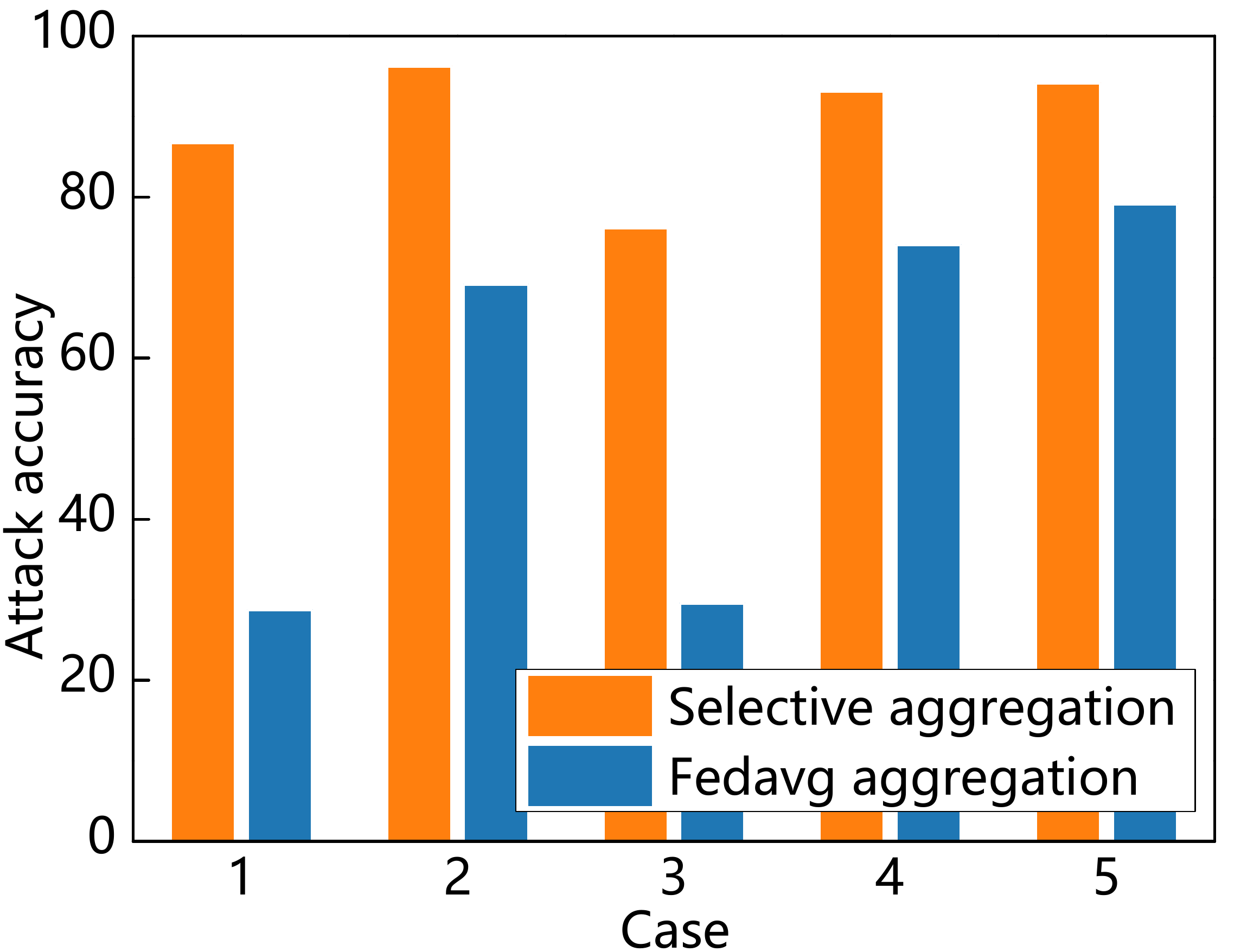}  
	\caption{The PPA's performance under common \texttt{Fedavg} aggregation and the presented selective aggregation.}  
	\label{aggregation}  
\end{figure} 

\begin{figure}
	\centering  
	\includegraphics [width = 0.3\textwidth]{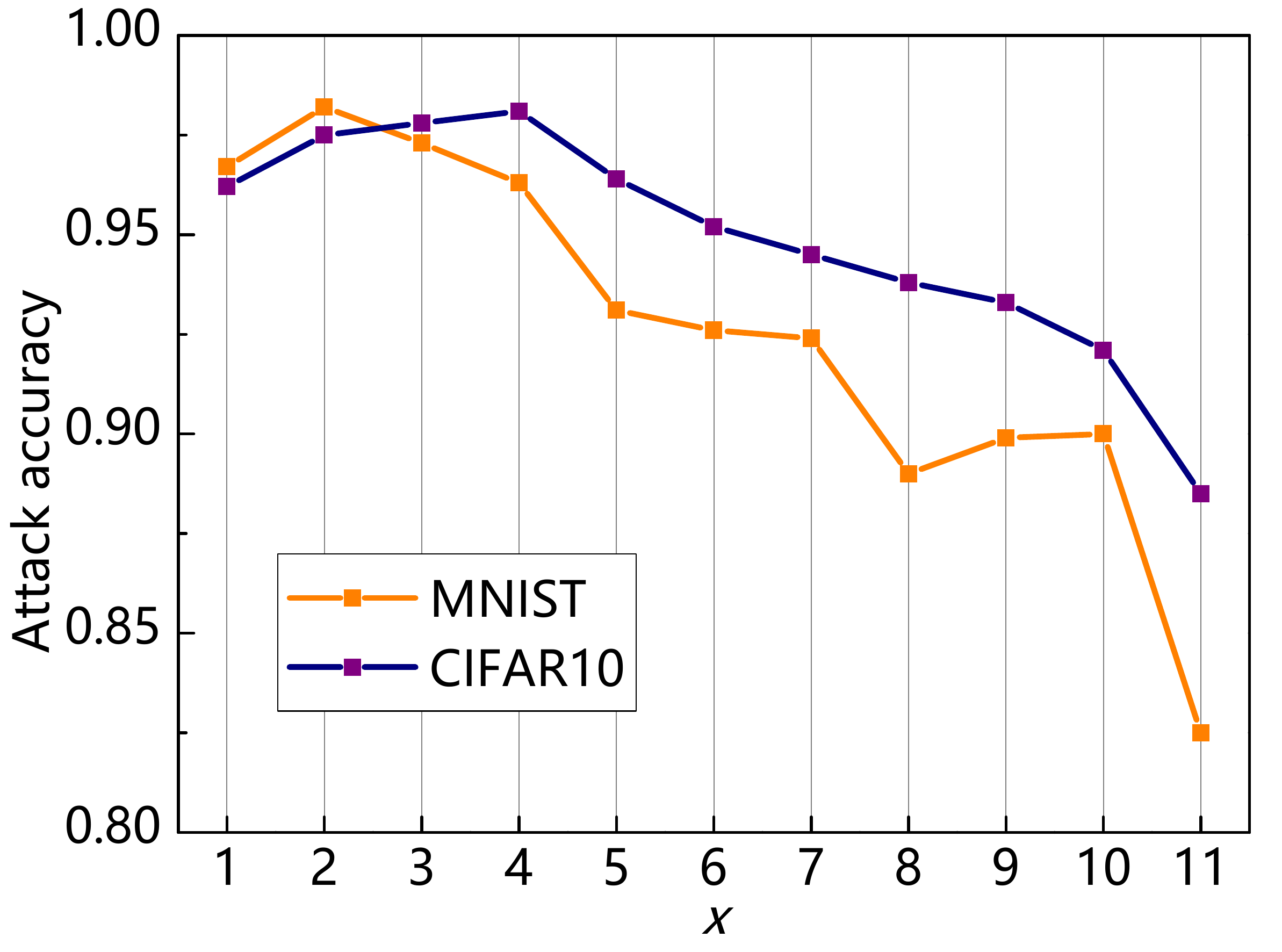}  
	\caption{Different $x$ values in selective aggregation.}  
	\label{k}  
\end{figure} 

\subsubsection{\textbf{Attack Performance Comparison}}
Figure~\ref{aggregation} indicates the accuracy comparison of PPA under default \texttt{FedAvg} aggregation \cite{McMahan2017Communication} and selective aggregation, respectively. We set up 5 cases under these two model aggregation methods. In the experiments, there are 10 users in FL, each user has 6,000 samples of MNIST dataset, the server has 150 samples per class, \textit{CP} and \textit{CD} are randomly selected from 40\% to 60\%, and $x$ = 4 in selective aggregation. The experimental results confirm that PPA has a higher accuracy under the selective aggregation compared to the default \texttt{FedAvg}, as the meta-classifier profiles the preference class according to $DS$. Compared with \texttt{FedAvg}, selective aggregation can improve $DS$ in each round and make the meta-classifier profile better.

\subsubsection{\textbf{Selection of Value $x$}}
Here we evaluate the PPA's performance as a function of $x$ that is the number of models chosen in the selective aggregation (recall Figure~\ref{Selective}). This experiment is conducted upon MNIST and CIFAR10 dataset, where the total number of users is 12 and each user has 4,000 samples. The PPA's attack accuracy with $x$ from 1 to 11 is depicted in Figure~\ref{k}. We can see that the attack is most effective when $x$ is $1$--$4$, and then the attack success rate decreases as the number of aggregated models increases. 
This is not difficult to understand, using an extreme case, when $x$ = 11, selective aggregation is equal to the default \texttt{FedAvg} aggregation, and the attack success rate largely depends on the data distribution of specific users in FL, as shown in Figure~\ref{aggregation}. In other words, the PPA's attack accuracy for a given user under \texttt{FedAvg} aggregation is difficult to control, as the user's model sensitivity is solely dependent on its local data characteristics, where the attacker has no other means to improve its model sensitivity opposite to the selective aggregation. The intuitive aggregation of models renders less preference sensitivity change, thus making the meta-classifier more difficult to accurately profile the preference. In this context, the value of $x$ should be properly selected according to the scale of users in FL in order to achieve optimal attacking performance.

\subsection{Real-World Case Studies}\label{sec:realworld}
Beyond the aforementioned MINST and CIFAR10 tasks, we leverage two more case studies to resemble real-world threats posed by PPA. In the first case, PPA attempts to profile popular sale items. In the second case, it infers the user emotions. Note here we further profile the top-$k$ preference classes, including majority and minority classes.
Because top-$k$ preference classes can already contain sufficient privacy information about user data distribution in real scenes. Consequently, we evaluate the top-$k$ output from the meta-classifier, and the results are summarized in Table~\ref{scene}. Experimental details are described below.

Note the top-$k$ accuracy here is relatively different from the one usually used in image classification tasks. The top-$k$ predicted classes are exactly the same as the top-$k$ ground-truth classes, but without caring the ranking order of these $k$ classes. For instance, the ground-truth preferred classes are $A, B, C, D, E$ in descending order. We regard that predicted cases of $\{ (A, B, C), (A, C, B), (B, A, C) \}$ in order are all acceptable and correct in our top-$k$ setting.

\begin{table}
\small
	\centering
	\caption{PPA in practical scenes.} 
	\label{scene}

\begin{tabular}{c|l|l|c}
\hline
Scene                                                                                  & Preference Class & \multicolumn{1}{c|}{PPA Output} & \begin{tabular}[c]{@{}c@{}}Attack \\ Accuracy\end{tabular} \\ \hline
\multirow{3}{*}{\begin{tabular}[c]{@{}c@{}} Shopping Scene\\ (Products-10K)\end{tabular}} & top1-(Food)        & top1-(Food)                       & 0.78                                      \\ \cline{2-4}
                                                                                       & \begin{tabular}[c]{@{}c@{}} top2-(Food,\\Clothing)\end{tabular}    & \begin{tabular}[c]{@{}c@{}} top2-(Food,\\Cosmetics)\end{tabular}      & 0.78                                                           \\ \cline{2-4}
                                                                                       & \begin{tabular}[c]{@{}c@{}} top3-(Food,\\Clothing,\\ Cosmetics)\end{tabular}   & \begin{tabular}[c]{@{}c@{}} top3-(Food,\\Cosmetics,\\ Clothing)\end{tabular}                   &    0.88                                                       \\ \hline
\multirow{2}{*}{\begin{tabular}[c]{@{}c@{}} Social Network\\ (RAF-DB)\end{tabular}}    & top1-(Disgusted)   & top1-(Disgusted)                  & 0.88                                 \\ \cline{2-4}
                                                                                       & \begin{tabular}[c]{@{}c@{}} top2-(Disgusted,\\Neutral)\end{tabular}     & \begin{tabular}[c]{@{}c@{}} top2-(Disgusted,\\Neutral)\end{tabular}                    & 1.00                                                           \\ \hline
\end{tabular}

\end{table}

\subsubsection{\textbf{Case I -- Customer Preference Profiler}}

We resemble an FL scenario based on shopping records of Products-10K dataset, which contains 10 shops, each with 4,000 shopping records. The \textit{CP} is 35\% and \textit{CD} is 32.5\% in the experiment, and the server has 1,500 auxiliary data samples in total. These shops are collaboratively training a global model. In the experiment, we randomly select a shop for PPA and profile its shopping preference (e.g., bestselling items can be a private concern). We extract the top-$k$ (i.e., $k = 2, 3$) preferences of its local dataset. The experimental results are shown in Table \ref{scene}. Among the shopping records of the customer we attack, the top-3 classes are Food, Clothing and Cosmetics. The final outputs of PPA are Food, Cosmetics and Clothing, and the attack accuracy is 78\%. It can be seen that without considering the top-3 order, PPA can infer customers' shopping habits with 78\% attack accuracy. Therefore, the server can successfully infer popular items, then push relevant advertisements or give this information to its competitor.

\subsubsection{\textbf{Case II -- User Psychological Profiler}}
In this experiment, the dataset of RAF-DB is used, and the \textit{CP} is 67.5\% and \textit{CD} is 62.2\%. The server intends to infer the personality hidden in the expression model published by users on social networks. For example, when PPA finds that photos of negative emotions dominate in the model uploaded by users, it can speculate that the user's personality may be negative. The experimental results are shown in Table \ref{scene}. In the photo gallery of the users we attack, the two most common expressions are disgusted and neutrality. Then we can infer that the target user's personality tends to be negative. When users normally use their photos for training in FL, PPA can potentially profile the psychological character.

\subsection{Scalability of PPA}
We further investigate  three crucial factors relevant to the scalability of PPA in FL. The first is the number of users or participants, where most previous experiments use value of ten. The second is the local training epochs, where all previous experiments use default one local training epoch. The third factor is the size of the public auxiliary samples reserved by the attacker to facilitate the model sensitivity extraction.
\subsubsection{\textbf{Number of Users}}
\begin{table}[]
\small
\caption{The PPA's performance under different user numbers.}
\label{user}
\centering
\begin{tabular}{c|c|cl}
\hline
\multirow{2}{*}{User Number} & \multirow{2}{*}{\begin{tabular}[c]{@{}c@{}}Attack \\ Accuracy\end{tabular}} & \multicolumn{2}{c}{Model Utility}                                     \\ \cline{3-4} 
                             &                                                                             & \multicolumn{1}{c|}{without Attack} & \multicolumn{1}{c}{with Attack} \\ \hline
10 users in FL               & 0.894                                                                       & \multicolumn{1}{c|}{0.758}          & \multicolumn{1}{c}{0.746}            \\ \hline
50 users in FL               & 0.820                                                                        & \multicolumn{1}{c|}{0.754}               &  \multicolumn{1}{c}{0.753}                                \\ \hline
100 users in FL              & 0.832                                                                       & \multicolumn{1}{c|}{0.753}               &   \multicolumn{1}{c}{0.751}                                \\ \hline
\end{tabular}
\end{table}

The number of users in FL we use to evaluate is 10, 50 and 100, respectively. Note that the user scale we considered, e.g., 100, is higher than existing orthogonal privacy inference attacks in FL~\cite{Hitaj2017GAN, Nasr2019Comprehensive,Melis2019Exploiting}, among which the highest user number that has been considered is 41 in~\cite{Hitaj2017GAN}. We conduct this experiment upon RAF-DB dataset, in which each user has 4,000 training data, \textit{CP} is 62.5\%, \textit{CD} is 57.5\%, and 100 auxiliary sets per class. 
Under this setting, we attack seven users who have different preference classes. The experimental results are shown in Table~\ref{user}. The result validates that the performance of PPA is insensitive to the user number, and the attack accuracy remains to be high, 83.2\%, in the case of 100 users. This is mainly because selective aggregation will aggregate the victim model with other user models that are most conducive to the attack, regardless of the number of users. In addition, we have affirmed that the model utility of global model in large-scale user scenarios is unaffected. More specifically, the model utility with PPA attack is similar to that of without PPA at various user scales (i.e., 50, 100), as shown in Table~\ref{user}.

\subsubsection{\textbf{Local Training Epochs}}\label{sec:localepoch}

\begin{table}[]
\small
\caption{The PPA's performance under different training epochs. }
\label{epoch}
\centering
\begin{tabular}{c|l|c}
\hline
\multicolumn{1}{l|}{Training Epoch} & PPA's Performance & \multicolumn{1}{l}{Average Accuracy} \\ \hline
\multirow{4}{*}{Epoch Num. = 1}            & User1: 0.92      & \multirow{4}{*}{0.88}                \\ \cline{2-2}
                                    & User2: 1     &                                      \\ \cline{2-2}
                                    & User3: 0.74      &                                      \\ \cline{2-2}
                                    & User4: 0.86      &                                      \\ \hline
\multirow{4}{*}{Epoch Num. = 20}           & User1: 0.88      & \multirow{4}{*}{0.87}                \\ \cline{2-2}
                                    & User2: 1     &                                      \\ \cline{2-2}
                                    & User3: 0.72      &                                      \\ \cline{2-2}
                                    & User4: 0.88      &                                      \\ \hline
\end{tabular}
\end{table}

\begin{figure*}
	\centering
	\subfigure[Model sensitivity of Class A and Class B.]{
		\includegraphics[width = 0.28\textwidth]{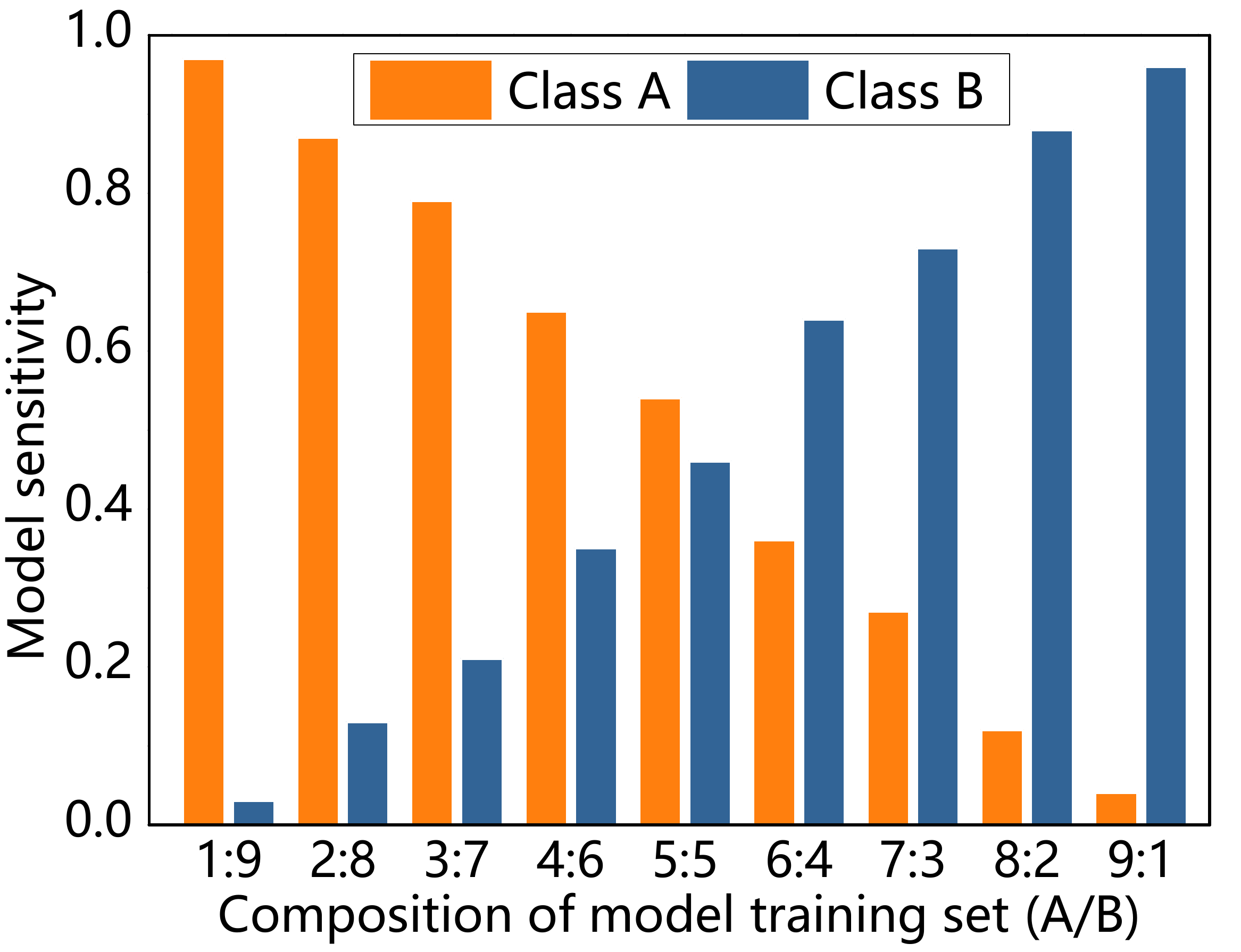}	
	}
	\subfigure[Model sensitivity of MNIST]{
		\includegraphics[width = 0.27\textwidth]{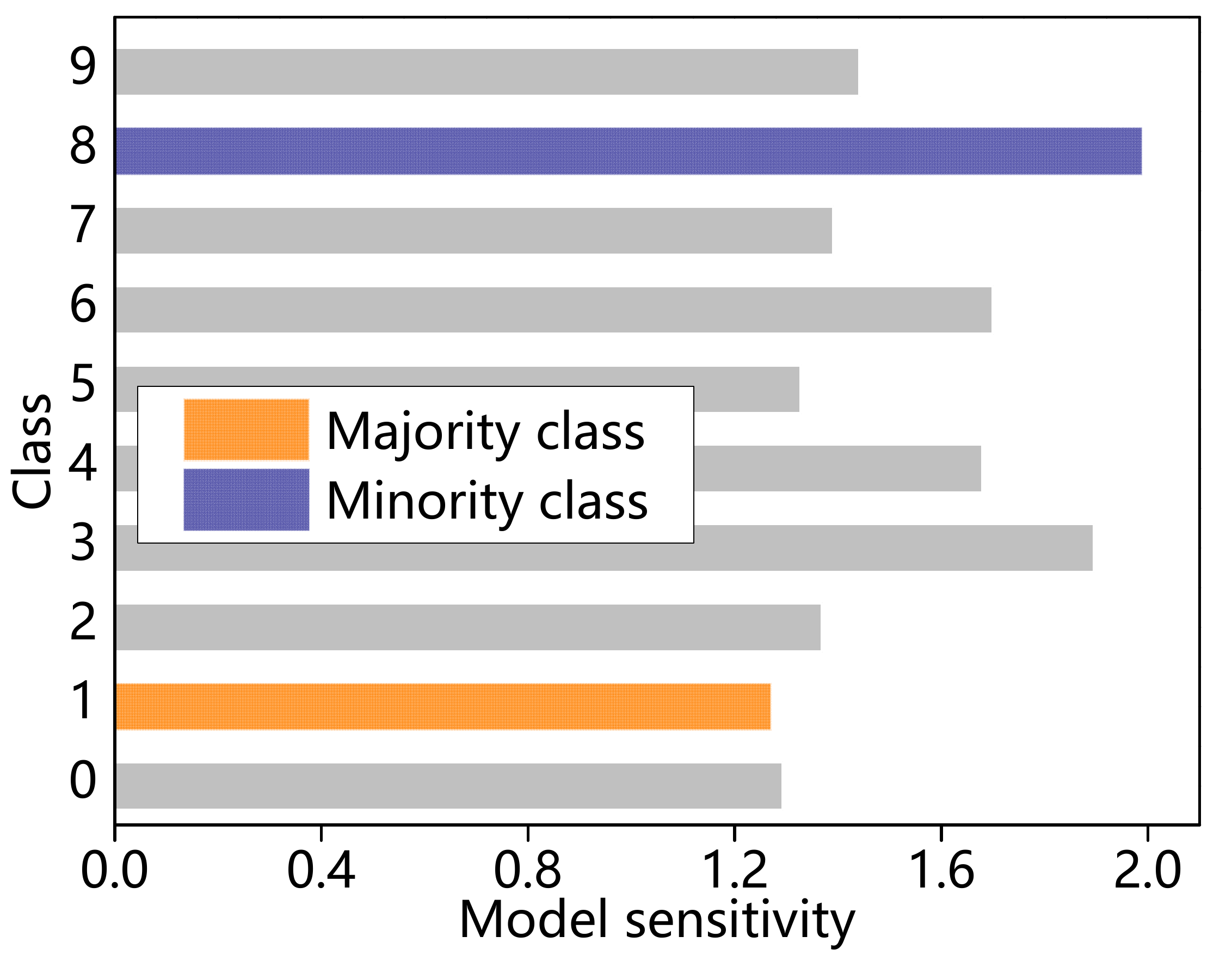}
	}
	\subfigure[Model sensitivity of CIFAR10]{
		\includegraphics[width = 0.3\textwidth]{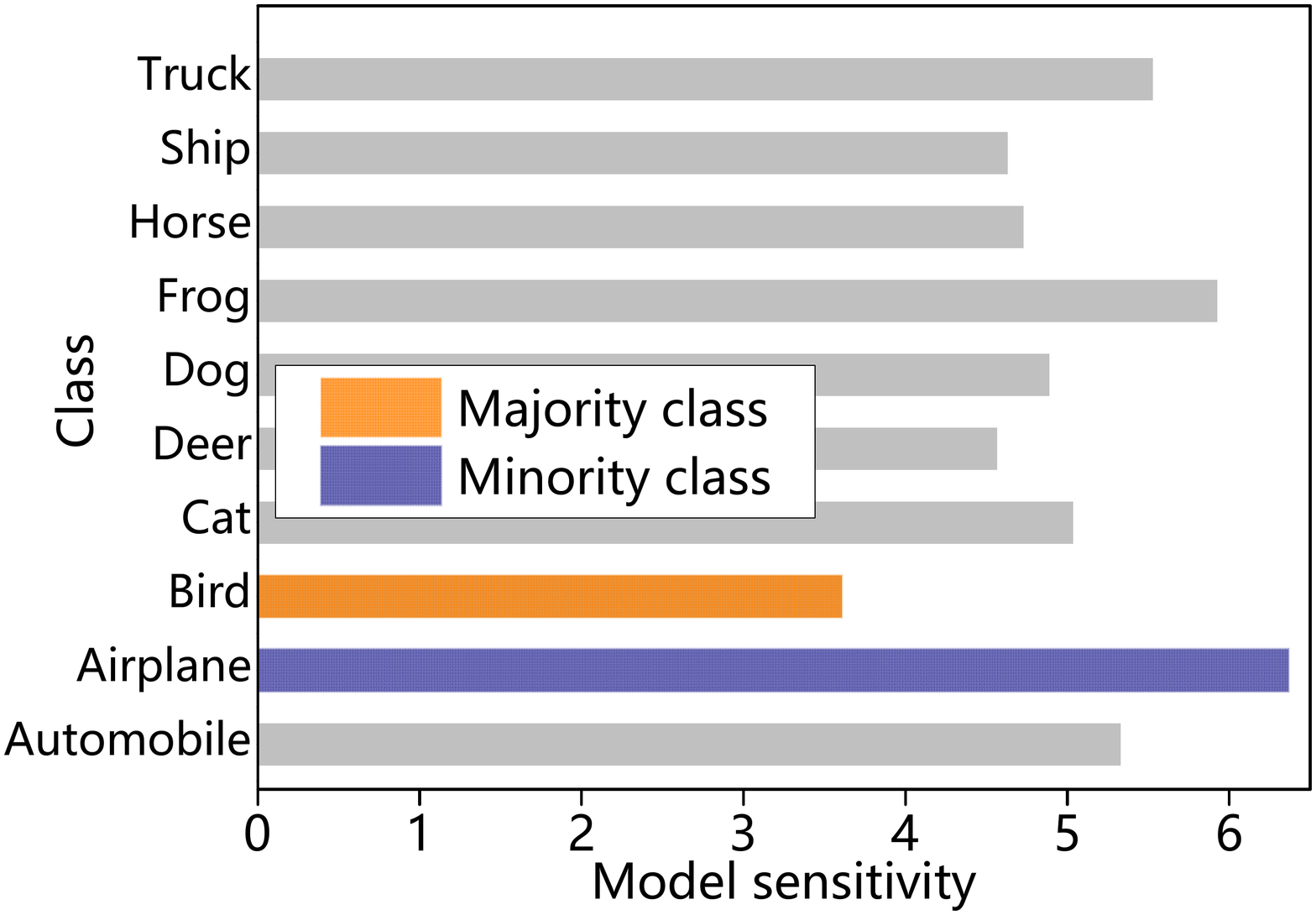}
	}		
	\caption{Model sensitivity change and extraction experiments.}
	\label{comparision}
\end{figure*}

A local user can control the number of local training epochs being more than 1 before updating the local model to the server. Here we evaluate the PPA's sensitivity to the the number of training epochs. 
In the experiments, RAF-DB is used as a training task and the auxiliary dataset is 150 per class. Each user has 3000 training data, and \textit{CP} is 62.5\% and \textit{CD} is 56\%. We first set training epoch to be 1, and randomly select 4 victim users to perform PPA. Then we increase the local training epochs to be 20. The PPA's performance under 1 local training epoch and 20 local training epochs are detailed in Table~\ref{epoch}. It can be observed from the experimental results that when users increase the number of local training epochs, the performance of PPA remains relatively stable, that is, the variance is slight without a clear tendency as the attack accuracy for user 4 increases. Thus, our empirical conclusion is that the effectiveness of PPA is independent of the epoch number of local training.

\begin{table}[]
\small
\caption{The PPA's performance under different numbers of auxiliary data. }
\label{aux}
\centering
\begin{tabular}{c|c}
\hline
Auxiliary Dataset & Average Accuracy \\ \hline
20 per class      & 0.675           \\ \hline
100 per class     & 0.819             \\ \hline
150 per class     & 0.86             \\ \hline
\end{tabular}
\end{table}

\subsubsection{\textbf{Auxiliary Data}}

The extraction of model sensitivity requires using a number of public auxiliary data. The more auxiliary data, the more comprehensive the sample features it contains, which will make the extracted model sensitivity close to the real situation. 
In this context, it is expected that the number of auxiliary data can affect the attack accuracy. We thus test the PPA's performance under different numbers of auxiliary data. The experimental setting is the same as that of local training epochs, and all auxiliary data in the experiment are randomly selected. We just vary the number of auxiliary data available to the server and evaluate the attack accuracy under 20, 100 and 150 auxiliary samples per class, which experimental results are shown in Table \ref{aux}. The results affirm that the more auxiliary sets, the better PPA's performance. Nonetheless, when there are few auxiliary sets (20 per class), our attack still has 67.5\% accuracy (the guessing accuracy is 14.3\%). According to our observation, PPA can perform well when the number of auxiliary sets is about 100 per class in our experiments.

\section{Discussion}
We further discuss the insights of PPA with more experiments. In addition, we demonstrate the effectiveness of PPA even when common privacy protection mechanisms (in particular, differential privacy and dropout) are applied by local users. We then evaluate the overhead of PPA, followed by demonstrating its practicality when attacking complicated models.

\begin{figure}
	\centering
	\subfigure[Majority sample]{
		\includegraphics[width=3.3cm]{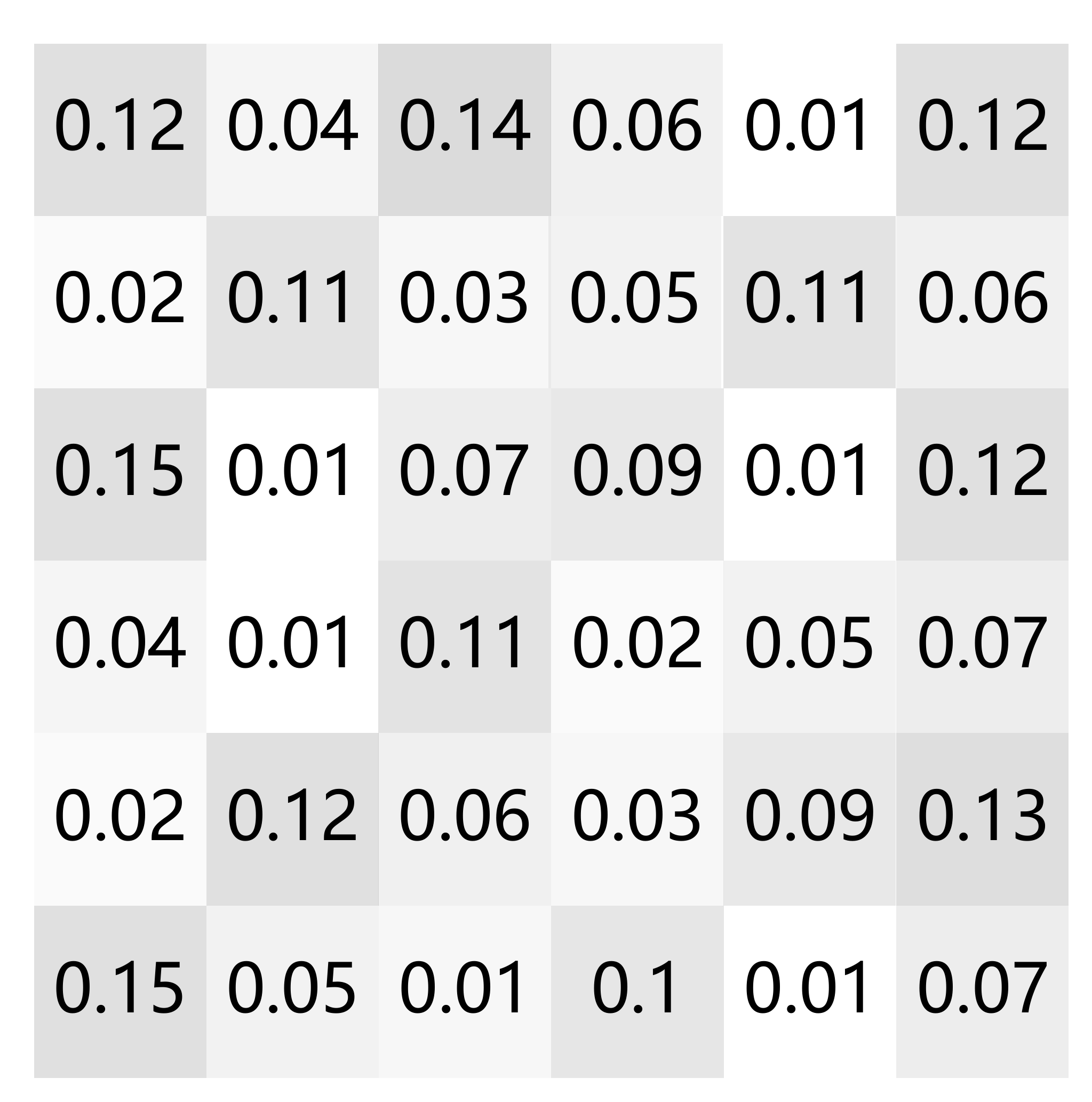}
		
	}
	\subfigure[Minority sample]{
		\includegraphics[width=3.3cm]{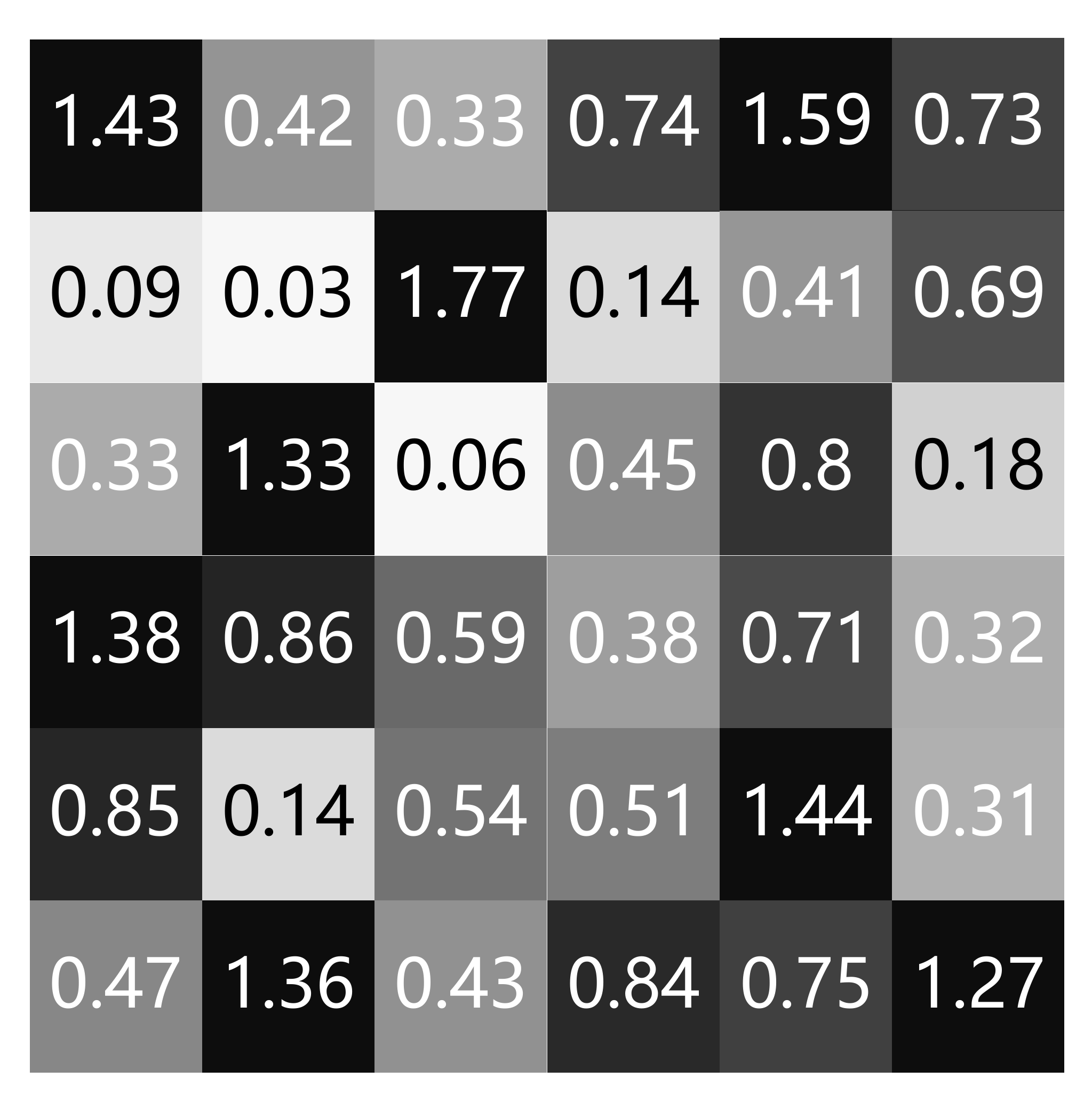}
	}
	\caption{The gradient change of 36 randomly selected neurons after retraining by samples of majority (a) and minority (b).}
	\label{gradient}
\end{figure}

\subsection{\textbf{The Rationale of PPA}} \label{sec:rationale}

Generally, we recognize that the user's model can remember the preference class, majority (minority) class, of the dataset during training, and reflects it \textit{in the form of gradient change}. The gradient value of neural network determines the convergence direction of the model, which is an important index to make the model tend to fit. We find that it can be exploited to infer whether the model is sensitive to certain data. To validate it, we have done a series of preliminary and detailed experiments, which are elaborated in the following.

To simplify the investigation, we focus on a binary classification task. We change the data distribution of the training set of binary classifiers based on MNIST dataset, and extract the gradient change when they are later retrained by two kinds of samples A and B, respectively. Specifically, we firstly adjust the proportion of A and B in training set from 1:9 gradually to 9:1, and consequentially train 9 binary classifiers, in which the total amount of training sets of each model is 6,000.
After model training, we retrain the model with A and B samples respectively, and record the model sensitivity of class A and B between the two models, as shown in Figure~\ref{comparision} (a).  The model sensitivity is linearly related to the  sample size of the given class.

Beyond the simplified binary classification, we also extract model sensitivities for each class in the MNIST and CIFAR10 models, as shown in Figure~\ref{comparision} (b) and Figure~\ref{comparision} (c). In this experiment, we train two models based on MNIST and CIFAR10 datasets, respectively. The training set is a random sample of 6,000 data to represent the data distribution of FL users. The majority class $\textit{1}$ and $Bird$ account for 50\%, minority class $\textit{8}$  and $Airplane$ accounts for 2\%, and the rest is evenly distributed. We affirm that the majority and minority classes correspond to the minimum and maximum values of the model sensitivity, respectively.

For further visualization, we extract the model sensitivity of majority and minority class in a model, by randomly selecting 36 neurons to show their gradient changes after retraining with a 6 $\times$ 6 square grayscale plot. The neurons are selected from the last convolutional layer of the model, as the gradients sensitivity is more salient in layers that are closer to the output than that of layers closer to the input due to the gradient decrease in the back-propagation. The larger the value of gradient change, the darker the color, as depicted in Figure~\ref{gradient}. We can see that the gradient change is much smaller retrained by the majority class samples (Figure~\ref{gradient} (a)) than that by the minority class samples (Figure~\ref{gradient} (b)). 
The gradient value of neural network determines the convergence direction of the model, which is an important index to direct the model to fit, and can essentially reflect whether the model is sensitive to certain data distribution characteristics in our PPA. 
More specifically, each sample in the training set contributes to minimizing the model loss through stochastic gradient descent algorithm. For the data point(s) that does not exist or have a small amount in the training set, the model does not have a good generalization, especially, at the beginning of the model training. Therefore, this data point(s) will have a greater gradient effect to change the weights of the corresponding neurons to minimize the expected loss of the model. As a matter of fact, this is also an important motivation for taking model aggregation in FL: when users have various data distributions, the server can still make the global model converge to all data categories by aggregating all the local models. Thus, the natural data heterogeneity characteristic in FL is vulnerable to the newly devised preference inference attack.

\subsection{\textbf{Defenses}}

Dropout~\cite{Dropout} and differential privacy~\cite{DifferentialPrivacy,DPSGD} are two widely used techniques in the fields of formal verification~\cite{WangDKZ20}, program synthesis~\cite{WangDXKZ21} and deep learning privacy protection~\cite{WuFSK20,TschantzSD20,Yu0PGT19}. Here we investigate the attack accuracy and model utility when they are adopted in FL that is under the attack of PPA. The experimental results detailed in Table~\ref{defence} demonstrate that the accuracy of PPA's meta-classifier drops when a smaller (i.e., the smaller, the stronger) privacy budget of $\epsilon$ of differential privacy is applied, and remains stable when the dropout is applied (i.e., CIFAR10). In the former case, even with a privacy budget as low as 1.21, our attacks still have a accuracy up to 57\% (guessing accuracy is 10\%). 


However, those protection mechanisms badly affect model utility or accuracy, as we have evaluated and shown in Table~\ref{defence}. It can be seen from the experimental results that the model utility (accuracy) degrades: a higher protection level (smaller $\epsilon$) renders a lower model utility. Though the dropout and differential privacy can mitigate the user preference leakage to some extent, they are with a notable undesired utility trade-off.


In addition, the client may check whether the global models distributed to other clients are the same, since the selective aggregation used by PPA sends different global models to different users. However, this check might not always be applicable. On one hand, a normal user is unlikely to see the global models sent to other users, as a standard secure communication channel between each user and the server is enforced to prevent a user's model from being exposed to other users. Otherwise, one malicious user can leak, for example  property privacy, of other users~\cite{Melis2019Exploiting}. On the other hand, recent personalized FL such as~\cite{PersonalizedFL} leverages a clustering technique to group users and indeed sends different global models to different groups of users by design. 

To significantly mitigate privacy  leakage in FL, cryptographic-based FL can be leveraged, where local models are aggregated in ciphertext. In this context, existing privacy inference attacks, such as~\cite{Shokri2017member,Nasr2019Comprehensive,Ganju2018Property,Melis2019Exploiting,Hitaj2017GAN}, and our proposed PPA cannot be immediately mounted. However, using a cryptographic approach will substantially increase the computational and communication overhead, preventing users with low computing resources and limited bandwidth from participating in FL.

\begin{table}  
\small
	\centering
	\caption{PPA attack accuracy and user model utility under defense settings. We experiment with Differentially-Private Stochastic Gradient Descent (DP-SGD) \cite{DPSGD}, the most representative DP mechanism for protecting machine learning models. We set $\epsilon_1$ = 96.9, $\epsilon_2$ = 19.38, $\epsilon_3$ = 4.85, $\epsilon_4$ = 1.21 under the constraints of noise multiplier = 0.05, 0.25, 1 and 4 respectively, and $\delta$ = $10^{-5}$.} 
	\label{defence}
\begin{tabular}{c|c|c|c|cccc}
\hline
\multirow{2}{*}{}                                                        & \multirow{2}{*}{Task} & \multirow{2}{*}{\begin{tabular}[c]{@{}c@{}}No\\ Def.\end{tabular}} & \multirow{2}{*}{Drop.} & \multicolumn{4}{c}{Differential Privacy}                                                 \\ \cline{5-8} 
                                                                         &                       &                                                                    &                        & \multicolumn{1}{c|}{$\epsilon_1$}   & \multicolumn{1}{c|}{$\epsilon_2$}   & \multicolumn{1}{c|}{$\epsilon_3$}   & $\epsilon_4$   \\ \hline
\multirow{2}{*}{\begin{tabular}[c]{@{}c@{}}Attack\\ Acc.\end{tabular}}   & MNIST                 & 0.89                                                               & 0.85                   & \multicolumn{1}{c|}{0.86} & \multicolumn{1}{c|}{0.75} & \multicolumn{1}{c|}{0.24} & 0.1  \\ \cline{2-8} 
                                                                         & CIFAR10                 & 0.97                                                               & 0.96                   & \multicolumn{1}{c|}{0.85} & \multicolumn{1}{c|}{0.79} & \multicolumn{1}{c|}{0.73} & 0.57 \\ \hline
\multirow{2}{*}{\begin{tabular}[c]{@{}c@{}}Model\\ Utility\end{tabular}} & MNIST                 & 0.90                                                               & 0.87                   & \multicolumn{1}{c|}{0.84} & \multicolumn{1}{c|}{0.81} & \multicolumn{1}{c|}{0.32} & 0.17 \\ \cline{2-8} 
                                                                         & CIFAR10                 & 0.80                                                               & 0.62                   & \multicolumn{1}{c|}{0.75} & \multicolumn{1}{c|}{0.64} & \multicolumn{1}{c|}{0.62} & 0.55 \\ \hline
\end{tabular}
\end{table} 

\begin{table}[]
\small
\caption{Computational overhead of PPA (time in seconds).} 
\label{overhead}
\begin{tabular}{cl|l|l|l|l}
\hline
\multicolumn{2}{c|}{Task} & \multicolumn{1}{c|}{MNIST} & \multicolumn{1}{c|}{CIFAR10} & \multicolumn{1}{c|}{\begin{tabular}[c]{@{}c@{}}Products\\ -10K\end{tabular}} & \multicolumn{1}{c}{\begin{tabular}[c]{@{}c@{}}RAF\\ -DB\end{tabular}} \\ \hline
\multicolumn{2}{c|}{\begin{tabular}[c]{@{}c@{}}Shadow model \\ training (offline)\end{tabular}}    & 839.29                & 2220.59               & 4725.21           & 3260.96       \\ \hline
\multicolumn{2}{c|}{\begin{tabular}[c]{@{}c@{}}Meta-classifier \\ training (offline)\end{tabular}} & 47.65                  & 38.44                  &    35.45         &  33.51      \\ \hline
\multicolumn{2}{c|}{\begin{tabular}[c]{@{}c@{}}Model sensitivity\\  extraction\end{tabular}}         & 27.19                 & 40.24                &  173.74           &    53.86    \\ \hline
\multicolumn{2}{c|}{Attack time}                                                                     & 0.37                  & 0.36                  &     0.34        &  0.27      \\ \hline
\multicolumn{2}{c|}{\begin{tabular}[c]{@{}c@{}}Selective\\  aggregation\end{tabular}}                                                           & 0.001                     & 0.006                     &   0.02          &   0.22     \\ \hline
\multicolumn{1}{c|}{\multirow{3}{*}{\begin{tabular}[c]{@{}c@{}}Local \\ training\end{tabular}}}                    & epoch=1                    &0.75 & 1.13&  2.07          &  3.35       \\ \cline{2-6} 
\multicolumn{1}{c|}{}                                                   & epoch=5                    &  1.52                     &       2.18                &    5.19         &  5.22      \\ \cline{2-6} 
\multicolumn{1}{c|}{}                                                   & epoch=10                   &    2.01                   &  3.46                     &   7.71& 9.13       \\ \hline
\end{tabular}
\end{table}

\subsection{\textbf{Computational Overhead}}
We have evaluated the computational overhead of the entire PPA attack life cycle, as detailed in the Table \ref{overhead}.
In this experiment, ten users are participated in FL, each owns 6,000 data. The auxiliary set is 200 samples per class, and the model learning rate is 0.001. 
The shadow model training time and meta-classifier training time dominate the computational overhead, because shadow models have to be established for each class. More accurate shadow model requires longer training time, while improving PPA online attack accuracy (as shown in Table VI). 
However, we should emphasize that these time-consuming preparations are carried out by the server in the \textit{offline} phase, without affecting the online training experience of FL when normal users participate the process. 
During online phase, the computational overhead of model sensitivity extraction is greater than that of other online operations. Firstly, the online model sensitivity extraction can be expedited by leveraging additional computation resources (e.g., parallelization) or future optimizations on the extraction process. 
Most importantly, this latency (waiting time) can be essentially disguised in real scene due to the inevitable heterogeneous computing resources and communication bandwidth available among FL users. There are always notable but normal latency caused by limited computation resource~\cite{Vepakomma2018Split} and communication bandwidth restrictions (e.g., uploading (downloading) the local (global) models)~\cite{Zhang0D21,MalandrinoC21a}. This normal latency will be prolonged when the number of users in FL is scaled up. Thereby, the users are expected to tolerate latency to a large extent.

\subsection{\textbf{Complicated Model}}\label{sec:complicated}
More complicated models can be vulnerable to PPA. Here we consider the VGG16. This experiment is based on the CIFAR10 dataset, where each user has 5,000 training data, CP is 60\%, CD is 57.5\%. The server has 800 samples per class as auxiliary data. The experimental results depict that PPA can achieve 88\% top-1 attack accuracy in average for all ten users. We observe that the averaged accuracy slightly drops about 8\% compared with a four layer convolutional network that is used. The potential reason is that the VGG16 has about 14,714,688 parameters in total (2,359,808 parameters in last convolutional layer accounting for a fraction of 16.04\%) , while the four layer model has 1,448,266 parameters (295,168 parameters in the last convolutional layer accounting for a fraction of 20.38\%). When the local model sensitivity is extracted, only the last convolutional layer is utilized. This means that a smaller fraction of model parameters is exploited in VGG16, which can result in degraded sensitivity, thus rendering a slightly lower profiling accuracy of PPA. The profiling accuracy can be expected to be improved for large models when a higher proportion of parameters are leveraged to extract the sensitivity. Nonetheless, relying on the sensitivity extracted from a single layer in a complicated model, in particular VGG16, PPA still achieves a high top-1 accuracy of 88\%, demonstrating its practicality of attacking large models. 


\section{Conclusion}
This work uncovered a new type of privacy inference attack (i.e., PPA) on FL to infer the preference classes of users. Based on our key observation that the gradient change reflects the sample size of a class in the dataset, we proposed techniques to quantify and extract such change as model sensitivity to profile the preference class assisted with the meta-classifier. Extensive experiments have shown that PPA is effective under various settings and poses real threats to preference-sensitive applications in real world, substantiated by two commercial applications. In addition, we have shown that PPA can be scalable in that it is insensitive to the number of FL participants and local training epochs employed by the user. While squarely using common privacy protection schemes can mitigate PPA, they introduce unacceptable model utility degradation.

\bibliographystyle{IEEEtranS}
\bibliography{PPA}
\end{document}